\documentclass[10pt,journal,compsoc,cspaper,final]{IEEEtran}
\ifCLASSOPTIONcompsoc
  \usepackage[nocompress]{cite}
\else
  \usepackage{cite}
\fi
\ifCLASSINFOpdf
\else
\fi

\usepackage{url}
\usepackage{mathtools} 
\usepackage{amsmath,amsfonts,amssymb}
\usepackage{algorithmic}
\usepackage{graphicx}
\usepackage{mathrsfs}
\usepackage{amsbsy}
\usepackage{amsthm}
\usepackage{booktabs}
\usepackage{textcomp}
\usepackage{bm}
\usepackage{ragged2e}

\usepackage{tikz}
\usepackage{tcolorbox}
\usetikzlibrary{arrows,shapes,chains}
\usetikzlibrary{spy}
\usepackage{diagbox} 
\usepackage{enumitem}
\usepackage{soul,xcolor} 
\usepackage[switch]{lineno} 
\usepackage{balance}
\usepackage{subfigure}
\usepackage{stfloats}
\usepackage{multirow}
\usepackage{xspace}
\usetikzlibrary{intersections}
\usepackage{caption}
\usepackage[nocompress]{cite}
\usepackage{algorithm}
\usepackage{makecell}
\usepackage{pifont}
\usepackage{bbding}
\usetikzlibrary{fit}
\usepackage{comment}
\usepackage{tabularx}
\usepackage[pagebackref,breaklinks,colorlinks]{hyperref}
\usepackage[export]{adjustbox}
\usepackage{boxedminipage}
\usepackage{colortbl}
\usepackage{pgfkeys}
\usepackage{flushend} 
\usepackage[capitalize]{cleveref}
\crefname{section}{Sec.}{Secs.}
\Crefname{section}{Section}{Sections}
\Crefname{table}{Table}{Tables}
\crefname{table}{Tab.}{Tabs.}

\newcommand{\etal}{\textit{et al}.}
\newcommand{\ie}{\textit{i}.\textit{e}.}
\newcommand{\eg}{\textit{e}.\textit{g}.}

\ifCLASSINFOpdf

\begin{document}


\author{Chi Zhang, Xiang Zhang, Mingyuan Lin, Cheng Li, Chu He, Wen Yang, Gui-Song Xia, Lei Yu
\IEEEcompsocitemizethanks{\IEEEcompsocthanksitem C. Zhang, M. Lin, C. He, W. Yang, and L. Yu are with the School of Electronic Information, Wuhan University, Wuhan 430072, China. E-mail: \{zhangchi1, linmingyuan, chuhe, yangwen, ly.wd\}@whu.edu.cn.
\IEEEcompsocthanksitem X. Zhang is with the Computer Vision Lab of ETH Zurich, Switzerland. E-mail: xiang.zhang@vision.ee.ethz.ch.
\IEEEcompsocthanksitem C. Li is with the Huawei Noah's Ark Lab, Shenzhen, China. E-mail: licheng89@huawei.com.
\IEEEcompsocthanksitem G. S. Xia is with the School of Computer Science, Wuhan University, Wuhan 430072, China. E-mail: guisong.xia@whu.edu.cn.
\IEEEcompsocthanksitem The research was partially supported by the National Natural Science Foundation of China under Grants 62271354, 62523111, and 61871297.
\IEEEcompsocthanksitem Corresponding author: Lei Yu.
}
}

\markboth{Submission to IEEE TPAMI}%
{Shell \MakeLowercase{\textit{et al.}}}

\title{CrossZoom: Simultaneously Motion Deblurring and Event Super-Resolving}

\IEEEtitleabstractindextext{
\begin{abstract}

\justifying
Even though the collaboration between traditional and neuromorphic event cameras brings prosperity to frame-event based vision applications, the performance is still confined by the resolution gap crossing two modalities in both spatial and temporal domains. This paper is devoted to bridging the gap by increasing the temporal resolution for images, \ie, motion deblurring, and the spatial resolution for events, \ie, event super-resolving, respectively. To this end, we introduce {\it C}ross{\it Z}oom, a novel unified neural {\it Net}work (CZ-Net) to jointly recover sharp latent sequences within the exposure period of a blurry input and the corresponding High-Resolution (HR) events. Specifically, we present a multi-scale blur-event fusion architecture that leverages the scale-variant properties and effectively fuses cross-modality information to achieve cross-enhancement. Attention-based adaptive enhancement and cross-interaction prediction modules are devised to alleviate the distortions inherent in Low-Resolution (LR) events and enhance the final results through the prior blur-event complementary information. Furthermore, we propose a new dataset containing HR {\it sharp-blurry} images and the corresponding {\it HR-LR} event streams to facilitate future research. Extensive qualitative and quantitative experiments on synthetic and real-world datasets demonstrate the effectiveness and robustness of the proposed method. Codes and datasets are released at \href{https://bestrivenzc.github.io/CZ-Net/}{https://bestrivenzc.github.io/CZ-Net/}.
\end{abstract}

}

\maketitle

\section{Introduction}
\def\imwidth{0.196}
\def\cimwid{0.1345}
\def\ccimwid{0.1310}
\def\rswidth{0.23}
\def\zuoxia{(-1.5,-1.15)}
\def\youshang{(-0.6,-0.45)}
\def\seqwidth{0.35}
\def\ssyy{(-0.8,-0.85)}
\def\ssizz{0.5cm}
\def\sssizz{1.8cm}
\def\sswidth{0.245\textwidth}
\def\ssmag{3}
\def\scc{(2.12,1.4)}
\def\boxseq{0.44}
\def\boxseqq{0.48}
\def\linewidthzz{2.8in}

\begin{figure}[t]
\footnotesize
	\centering

    \begin{minipage}[t]{\linewidth}
    	\centering
          \begin{figure}[H]
             \includegraphics[width=\linewidth]{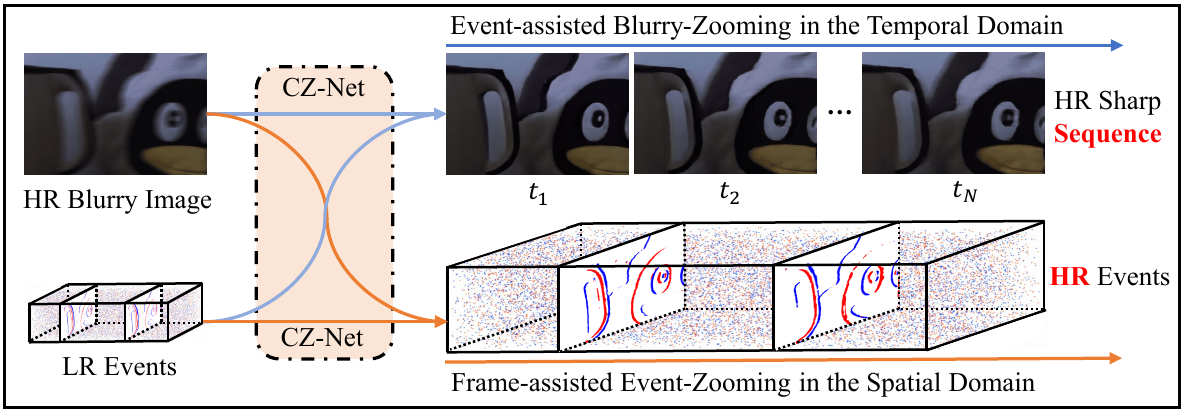}
          \end{figure}
    \end{minipage}%
    \\
    (a) CrossZoom Network
    \\
    \vspace{.4mm}
    \begin{minipage}[t]{\linewidth}
    		\centering
     \begin{tikzpicture}[inner sep=0]
            \node[
            label= {[label distance=-0.20cm,text depth=2.3ex,rotate=90,align=center] left:\textcolor{black}{\footnotesize {(b)}}}]{\includegraphics[width=\cimwid\linewidth,trim={0 0 0 0},clip]{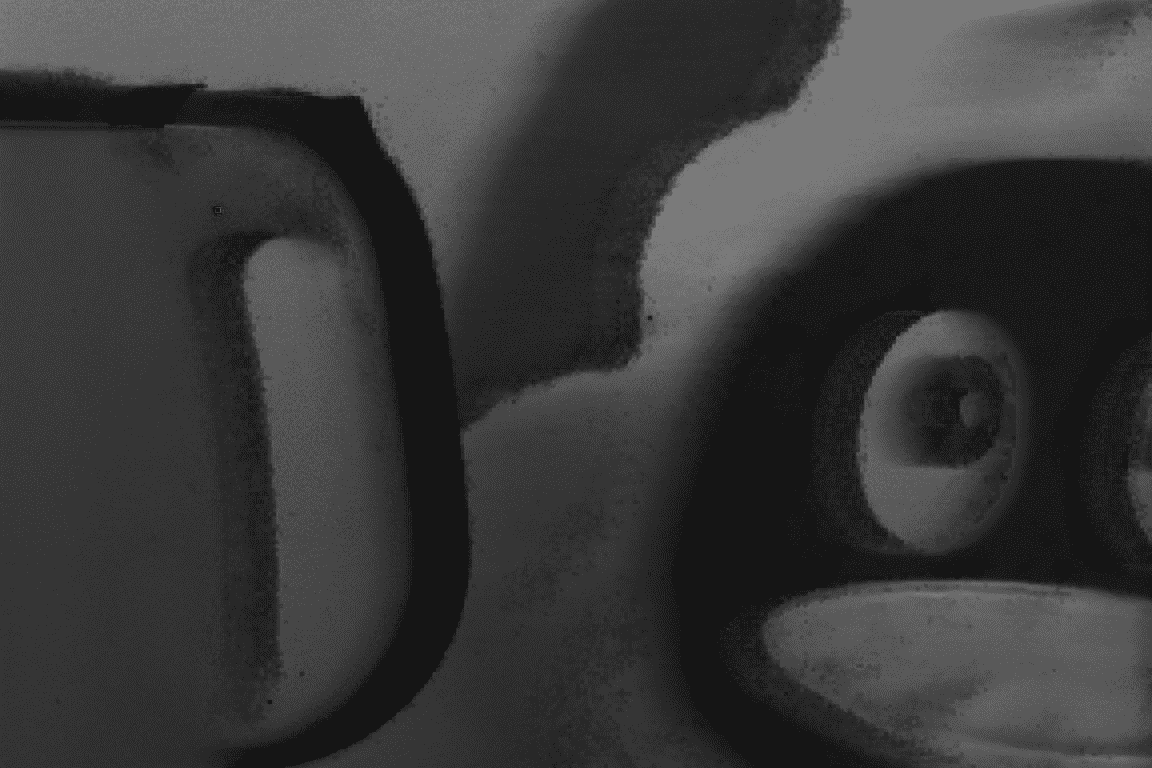}};
            \end{tikzpicture}\hfill
			\includegraphics[width=\cimwid\linewidth,trim={0 0 0 0},clip]{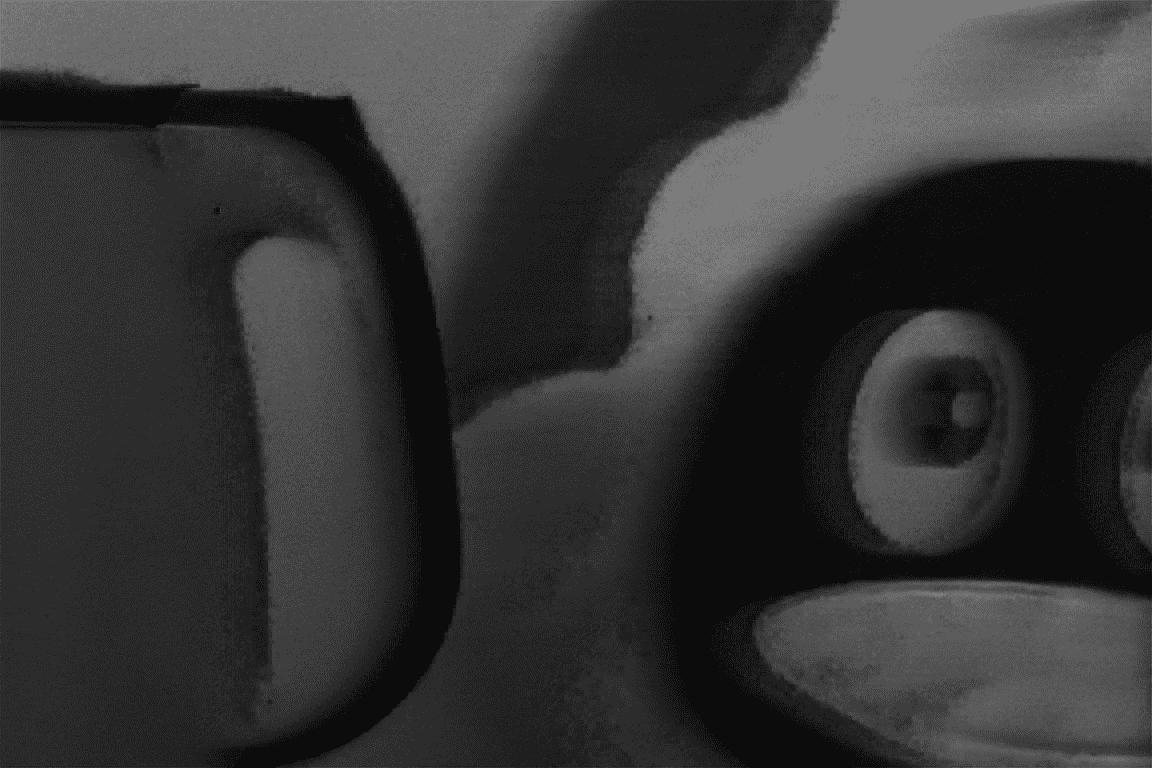}\hfill
			\includegraphics[width=\cimwid\linewidth,trim={0 0 0 0},clip]{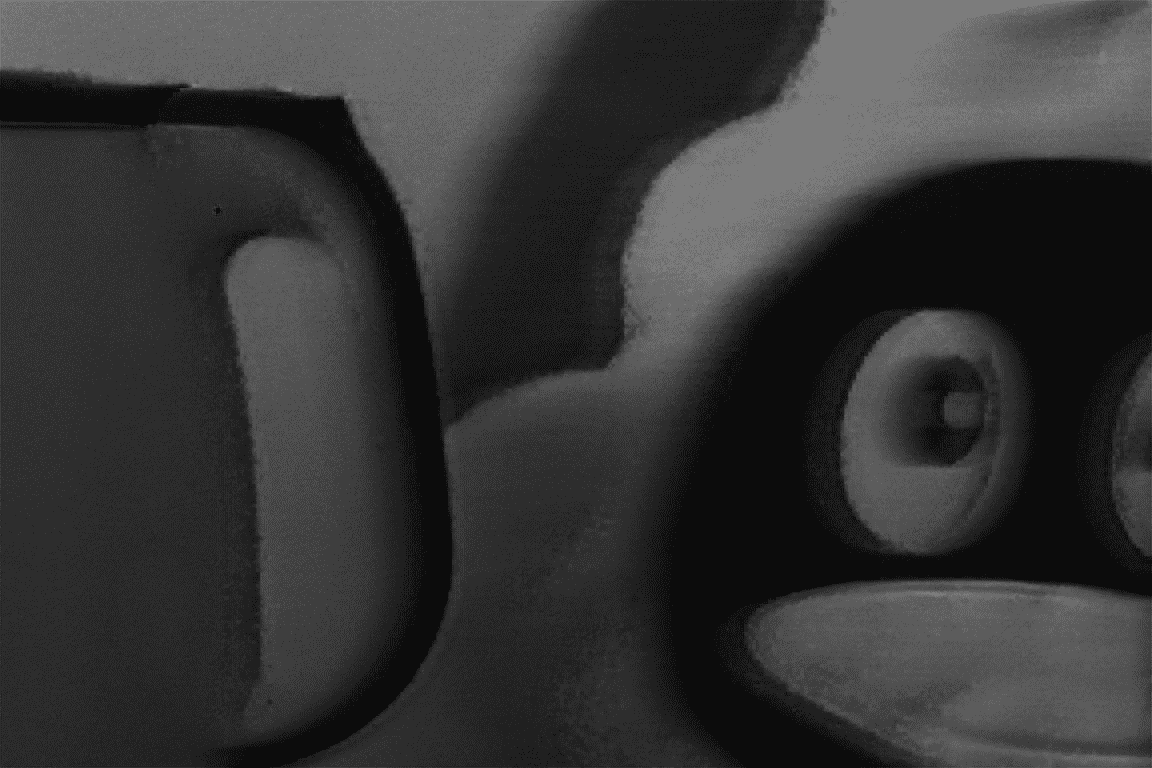}\hfill
			\includegraphics[width=\cimwid\linewidth,trim={0 0 0 0},clip]{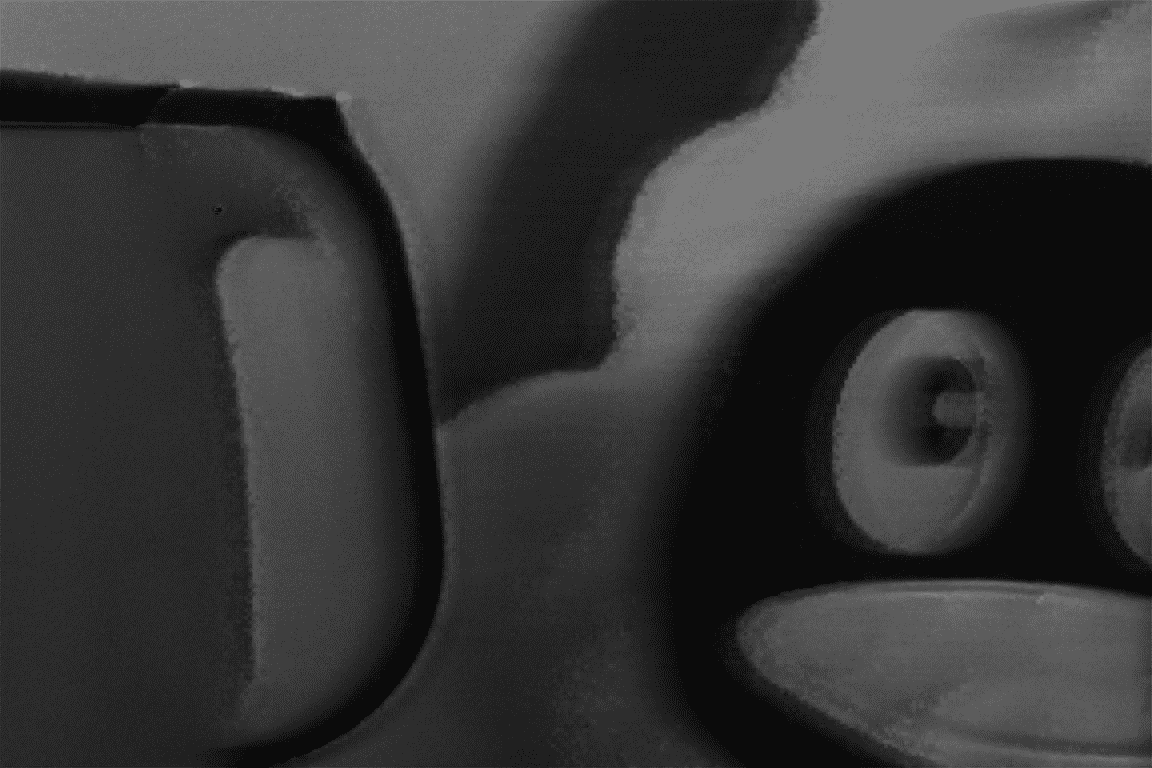}\hfill
			\includegraphics[width=\cimwid\linewidth,trim={0 0 0 0},clip]{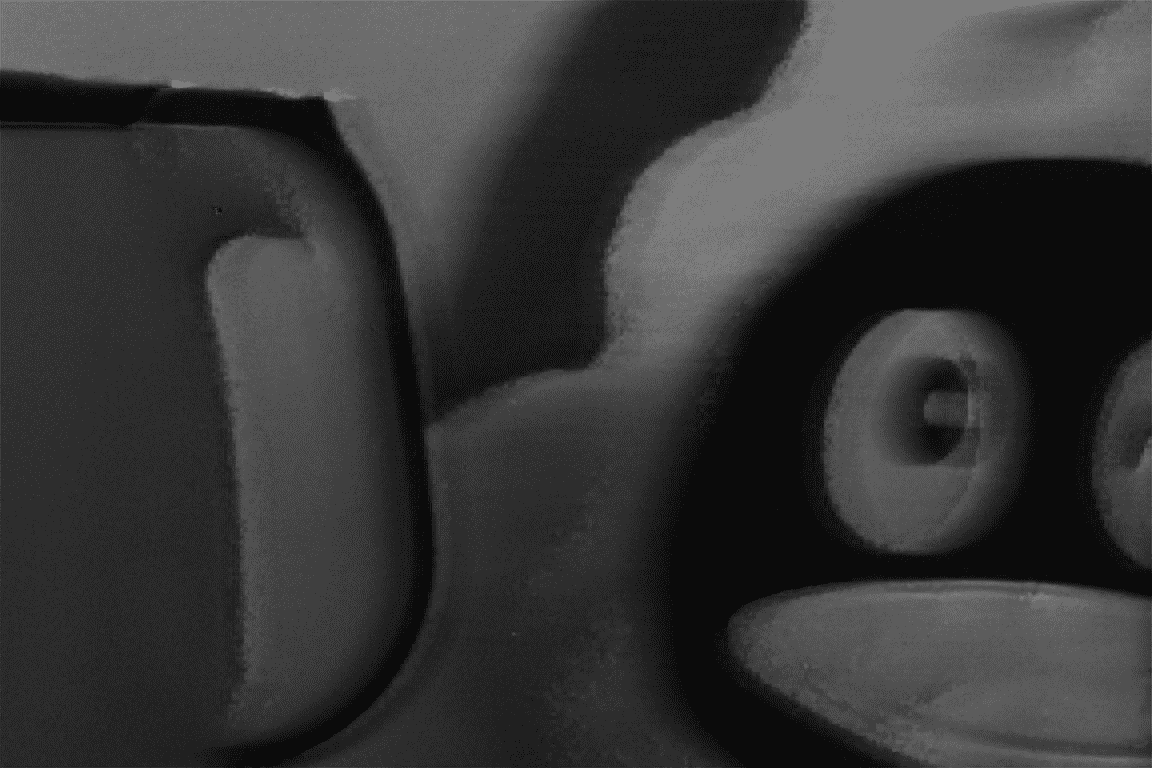}\hfill
			\includegraphics[width=\cimwid\linewidth,trim={0 0 0 0},clip]{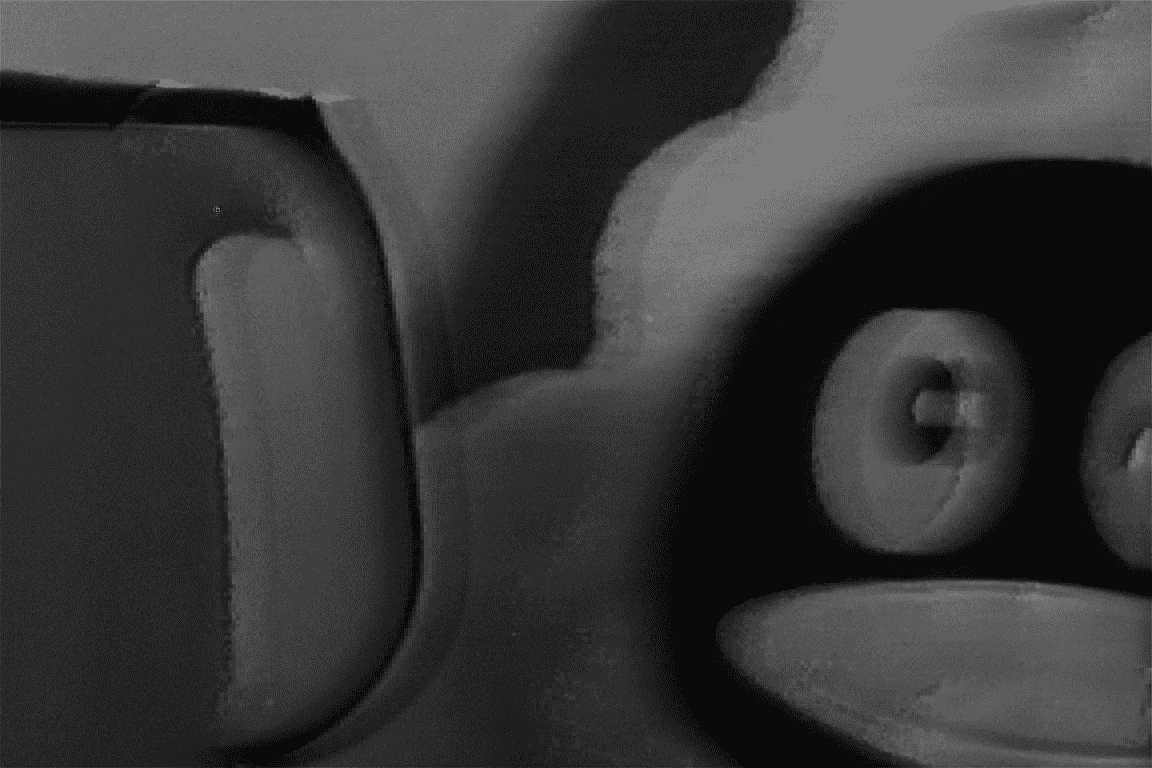}\hfill
			\includegraphics[width=\cimwid\linewidth,trim={0 0 0 0},clip]{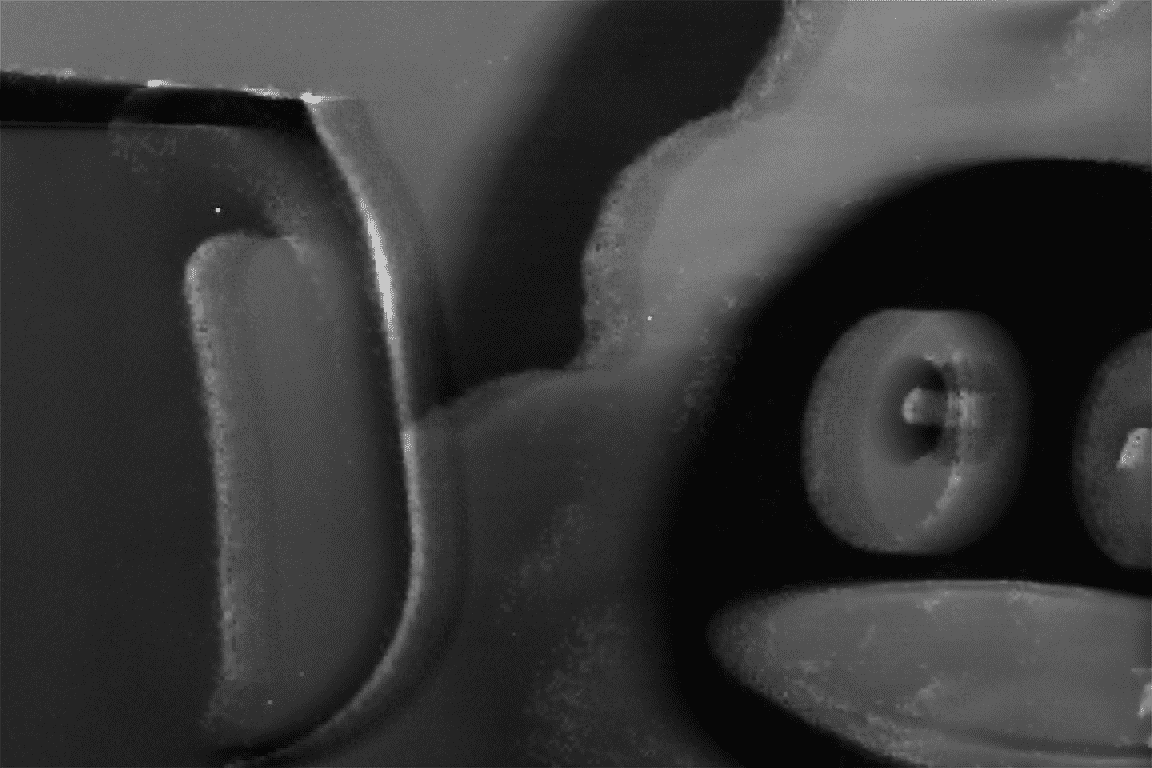}\hfill
    \end{minipage}
 \\
  \vspace{.2mm}
\begin{minipage}[t]{\linewidth}
    		\centering
      \begin{tikzpicture}[inner sep=0]
            \node[
            label= {[label distance=-0.20cm,text depth=2.3ex,rotate=90,align=center] left:\textcolor{black}{\footnotesize {(c)}}}]{\includegraphics[width=\cimwid\linewidth,trim={0 0 0 0},clip]{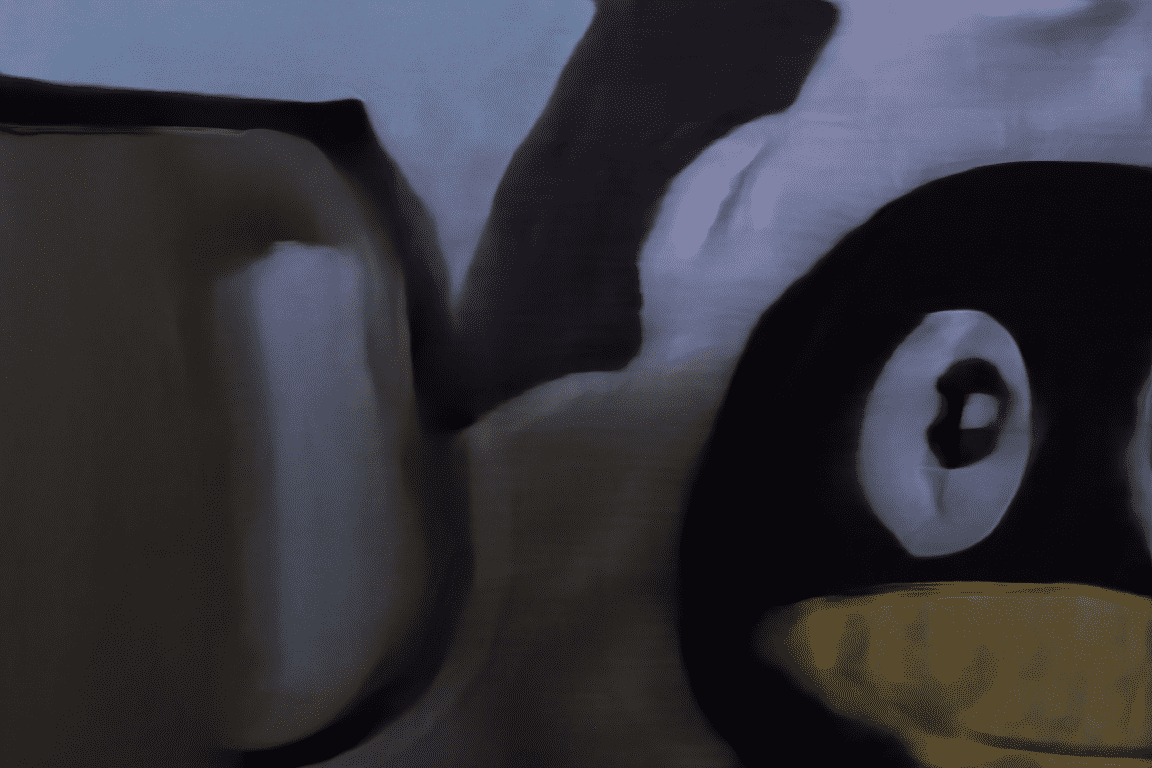}};
            \end{tikzpicture}\hfill
			\includegraphics[width=\cimwid\linewidth,trim={0 0 0 0},clip]{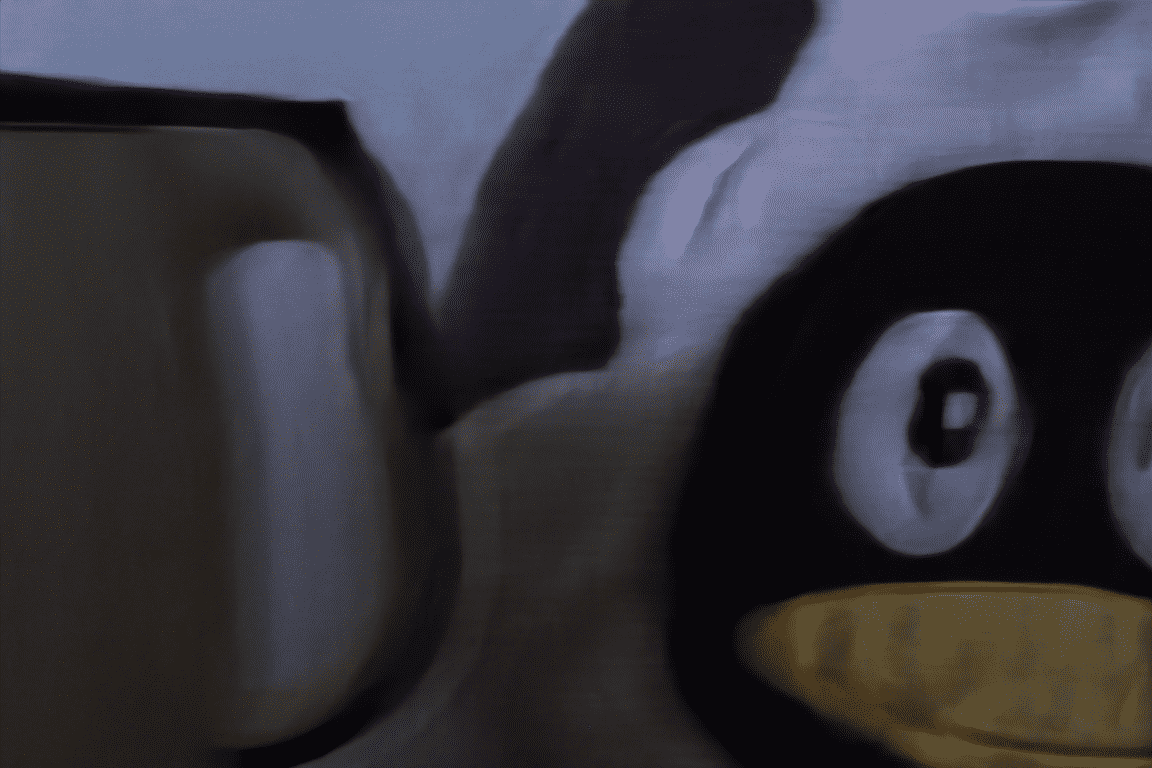}\hfill
			\includegraphics[width=\cimwid\linewidth,trim={0 0 0 0},clip]{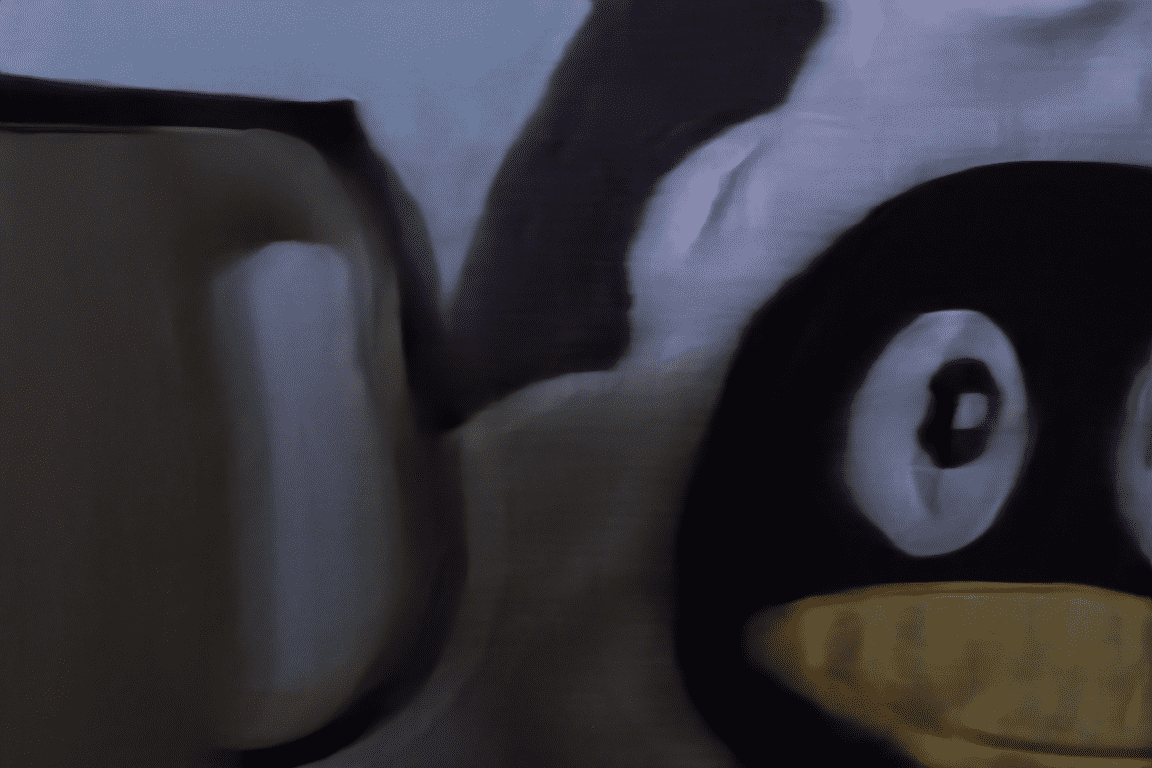}\hfill
			\includegraphics[width=\cimwid\linewidth,trim={0 0 0 0},clip]{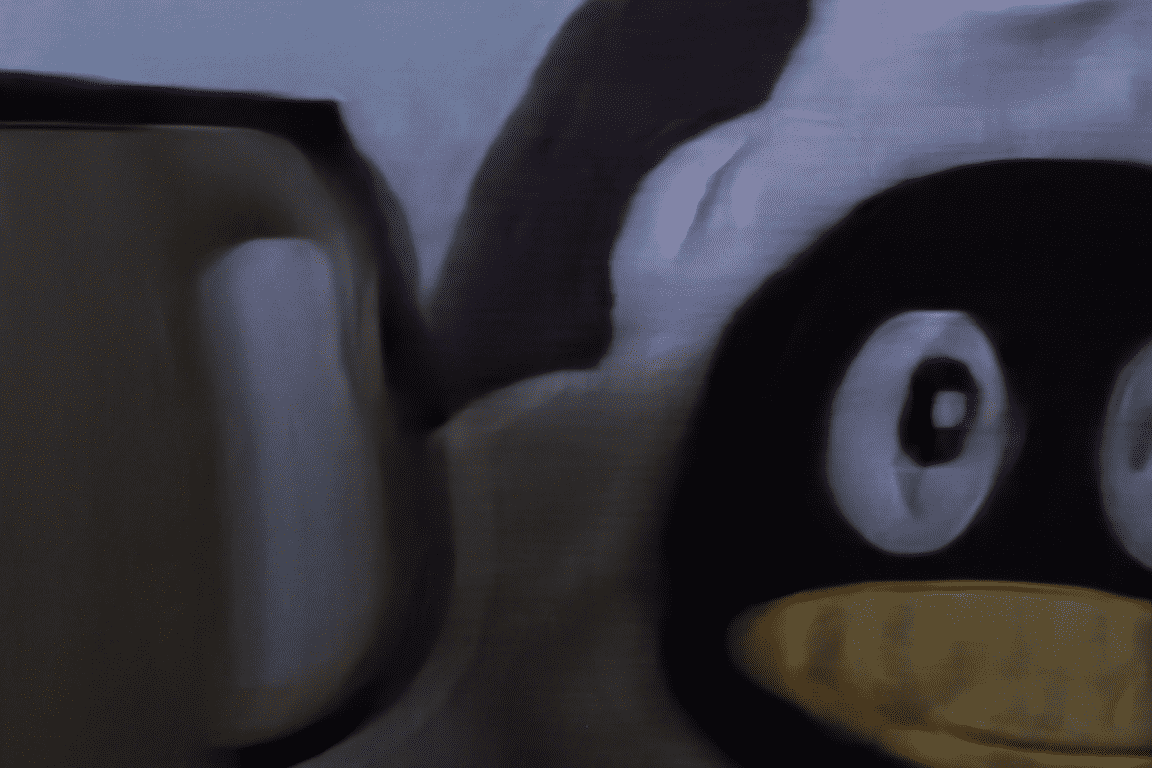}\hfill
			\includegraphics[width=\cimwid\linewidth,trim={0 0 0 0},clip]{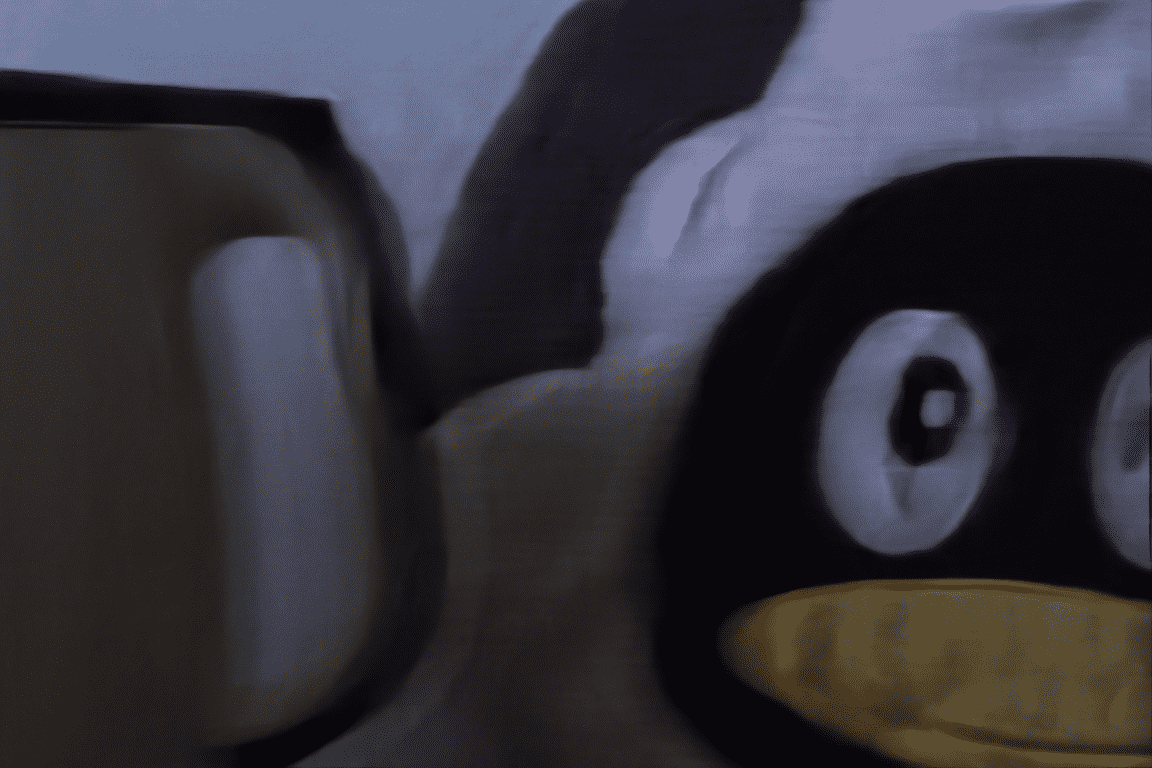}\hfill
			\includegraphics[width=\cimwid\linewidth,trim={0 0 0 0},clip]{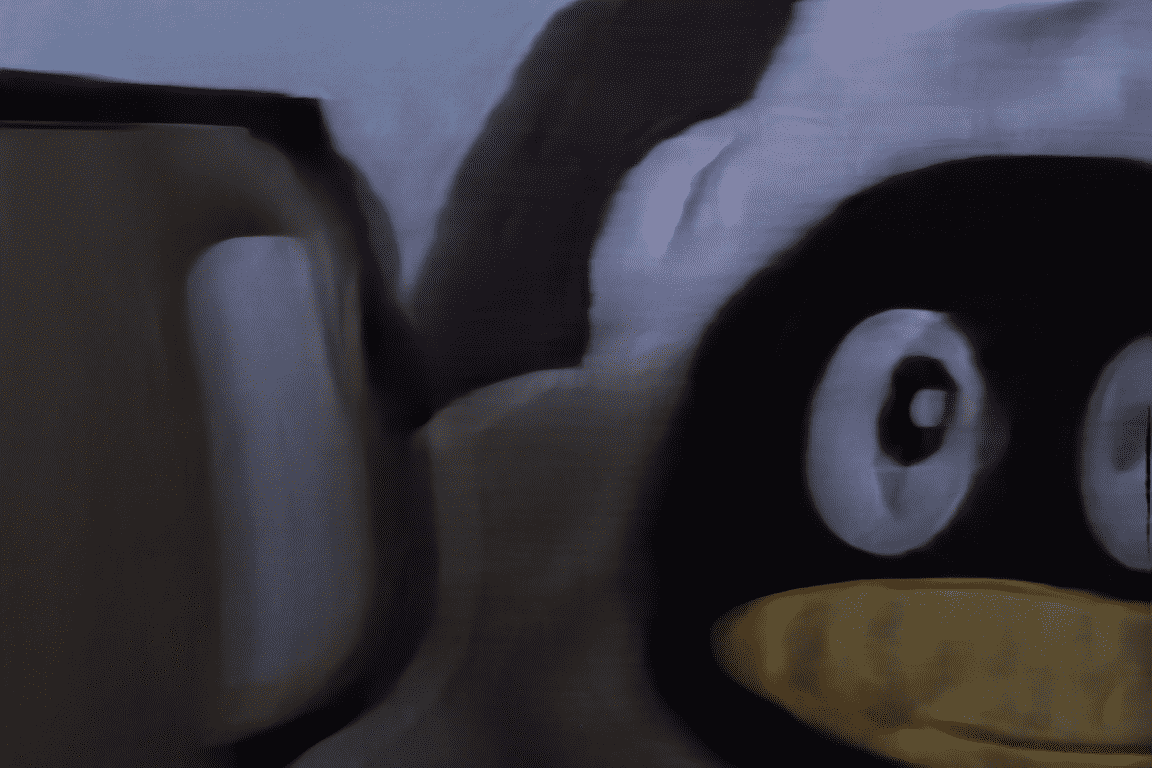}\hfill
			\includegraphics[width=\cimwid\linewidth,trim={0 0 0 0},clip]{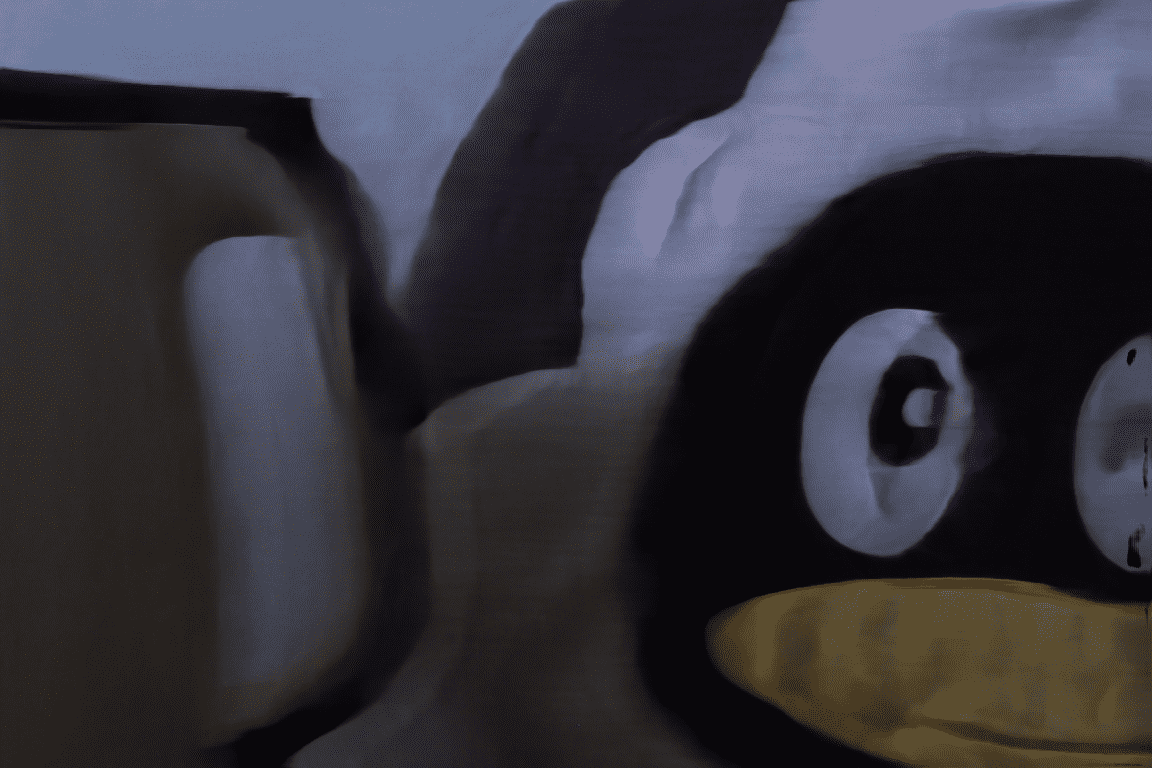}\hfill
    \end{minipage}
    \\
     \vspace{.2mm}
\begin{minipage}[t]{\linewidth}
    		\centering
      \begin{tikzpicture}[inner sep=0]
            \node[
            label= {[label distance=-0.20cm,text depth=2.3ex,rotate=90,align=center] left:\textcolor{black}{\footnotesize {(d)}}}]{\includegraphics[width=\cimwid\linewidth,trim={0 0 0 0},clip]{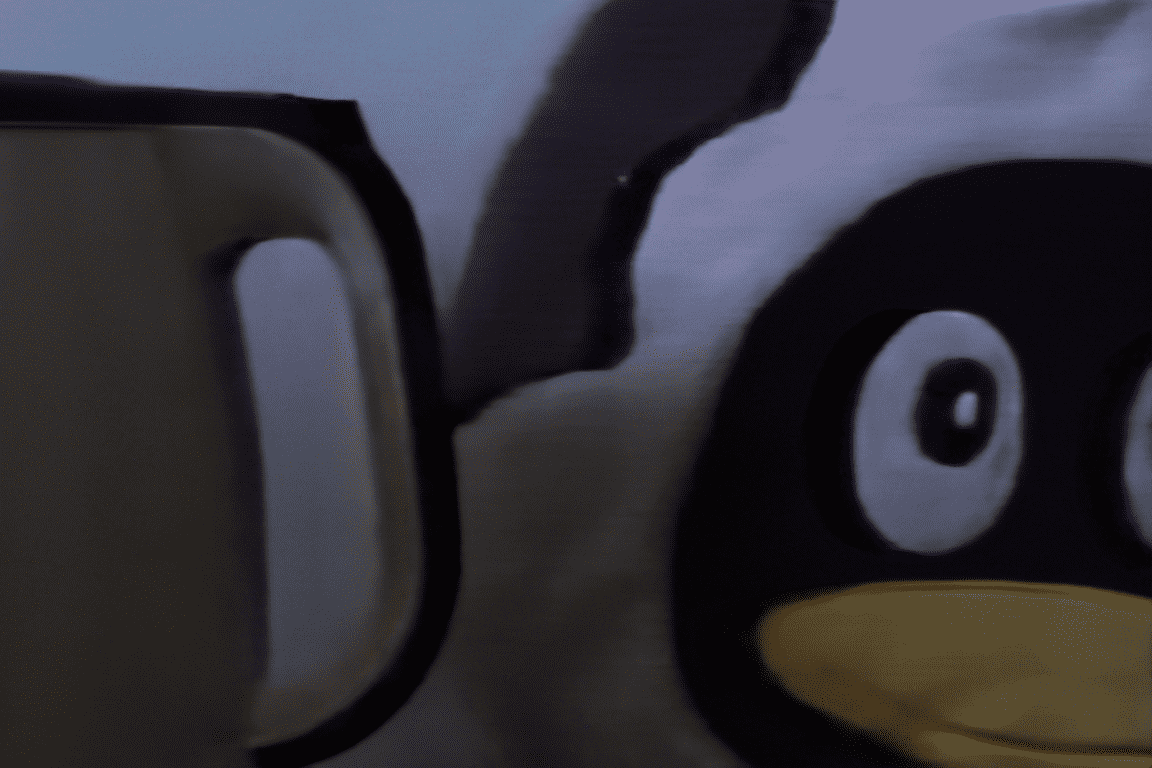}};
            \end{tikzpicture}\hfill
			\includegraphics[width=\cimwid\linewidth,trim={0 0 0 0},clip]{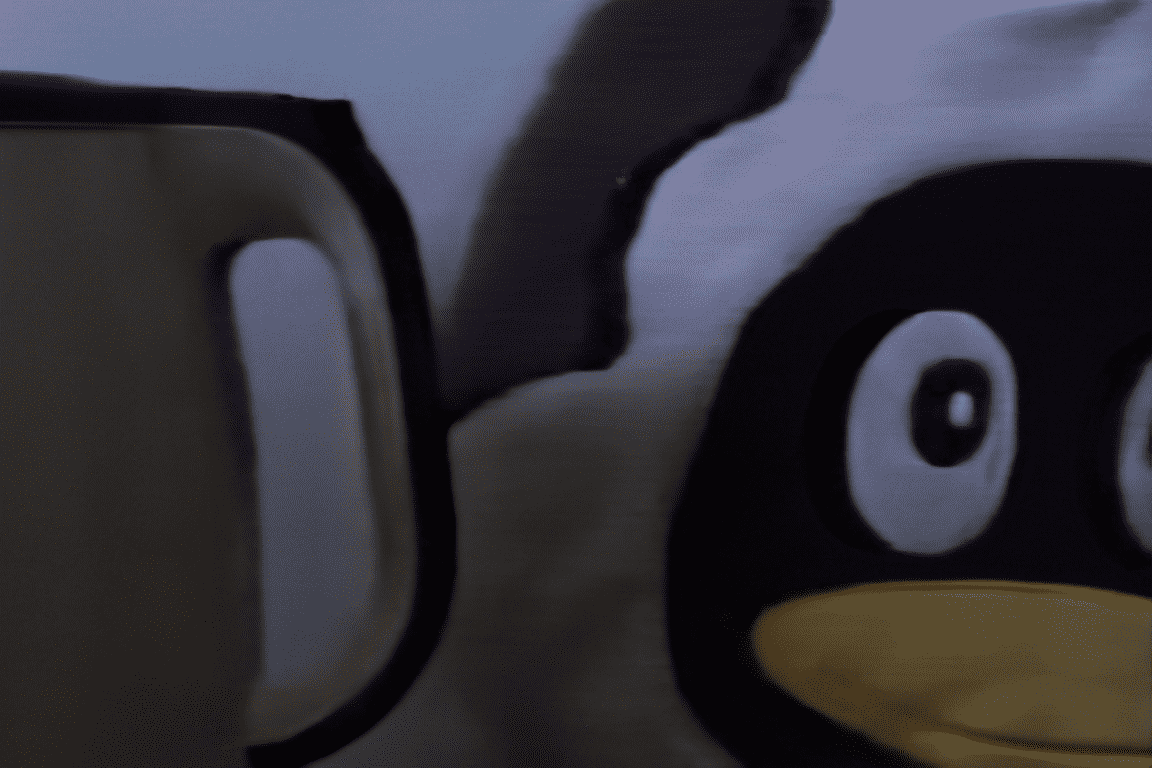}\hfill
			\includegraphics[width=\cimwid\linewidth,trim={0 0 0 0},clip]{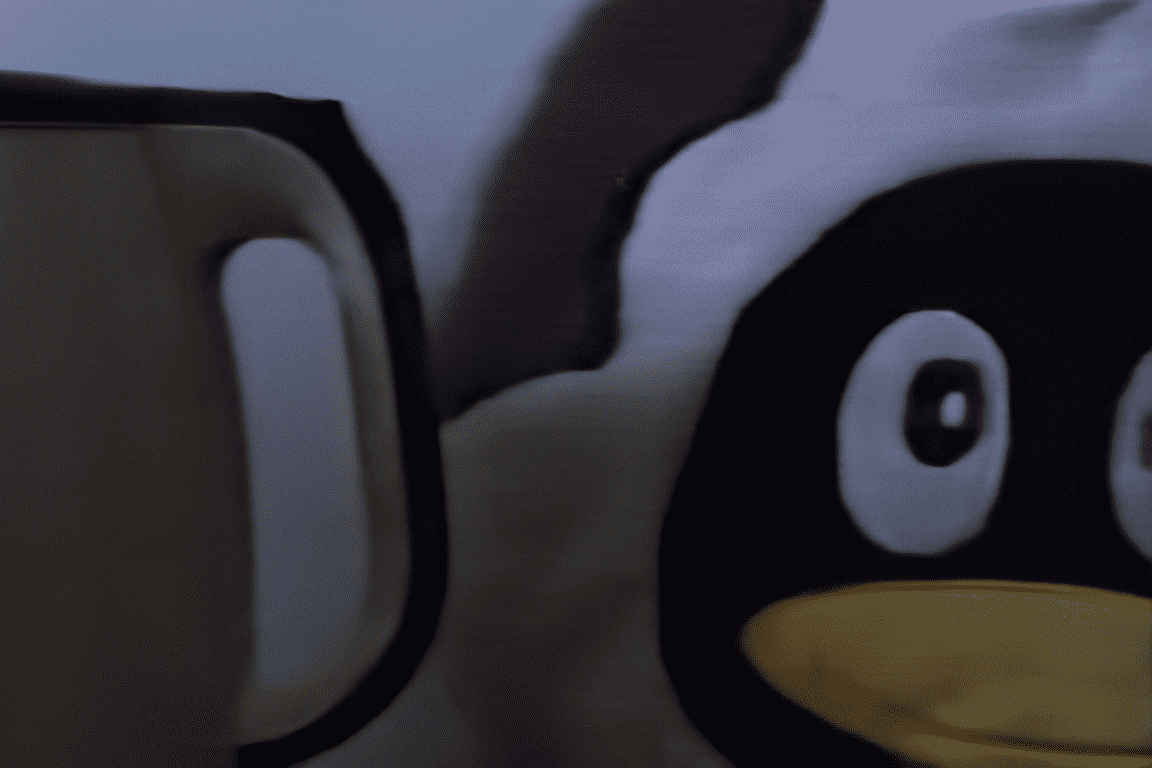}\hfill
			\includegraphics[width=\cimwid\linewidth,trim={0 0 0 0},clip]{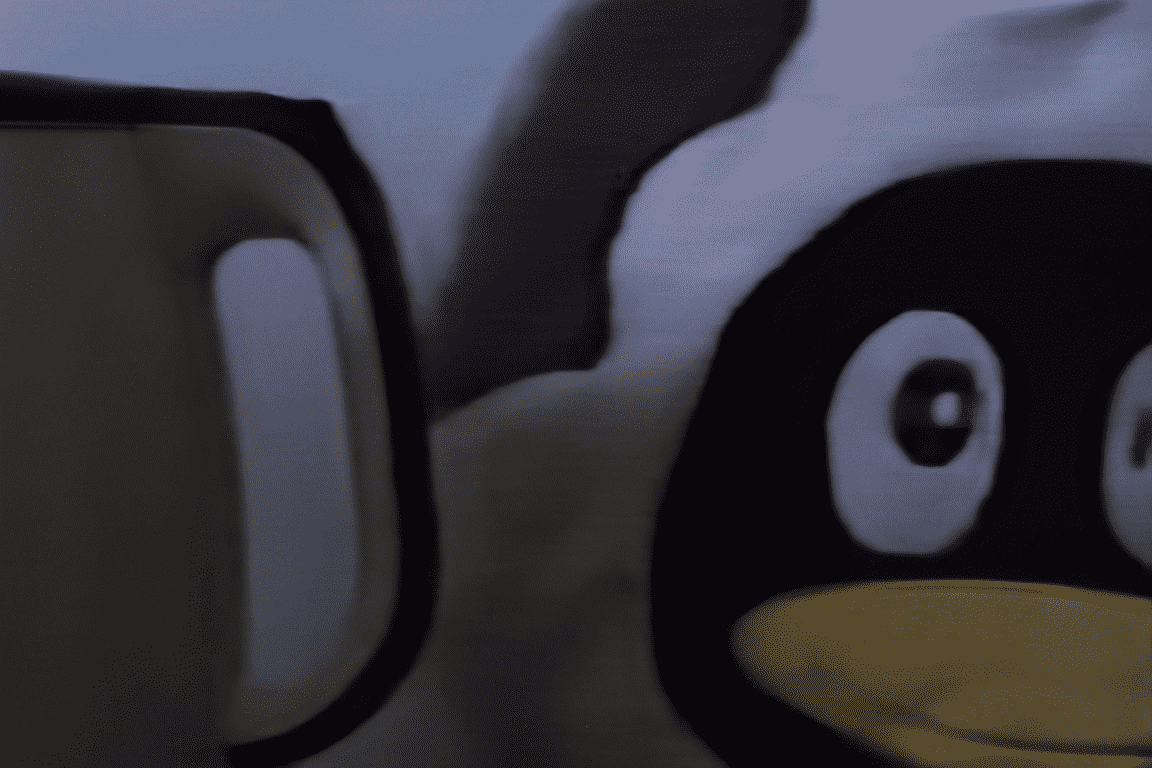}\hfill
			\includegraphics[width=\cimwid\linewidth,trim={0 0 0 0},clip]{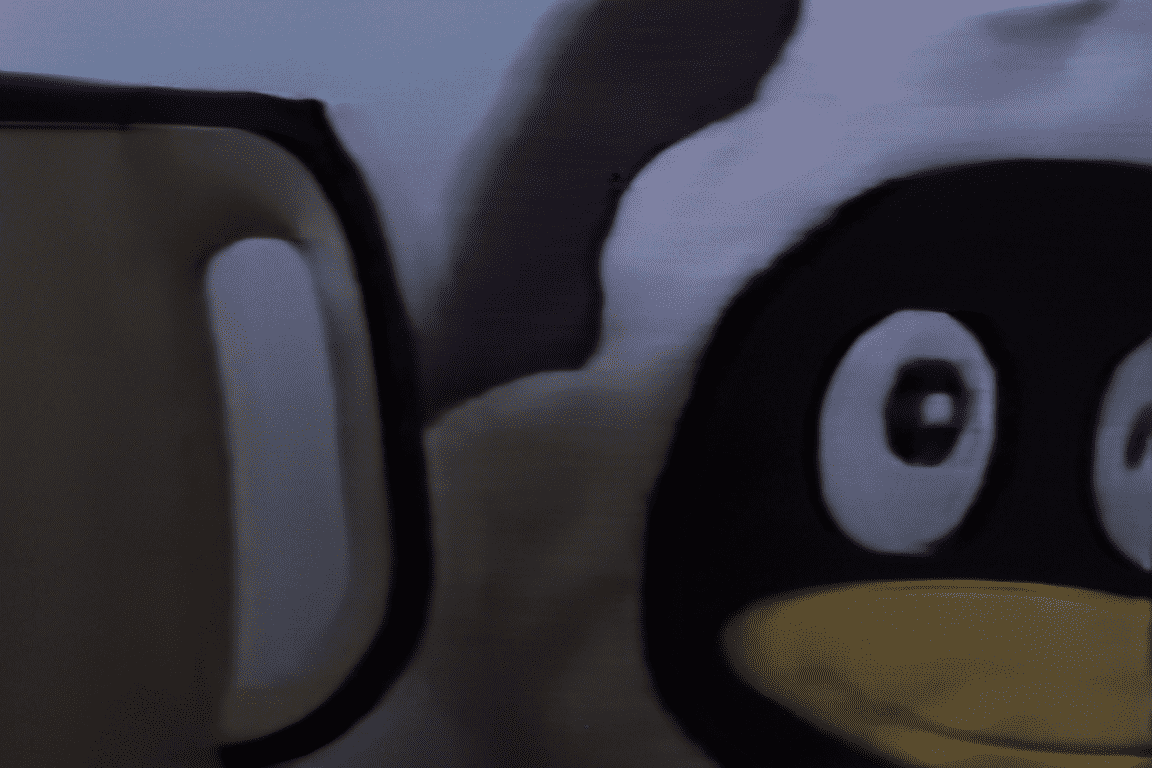}\hfill
			\includegraphics[width=\cimwid\linewidth,trim={0 0 0 0},clip]{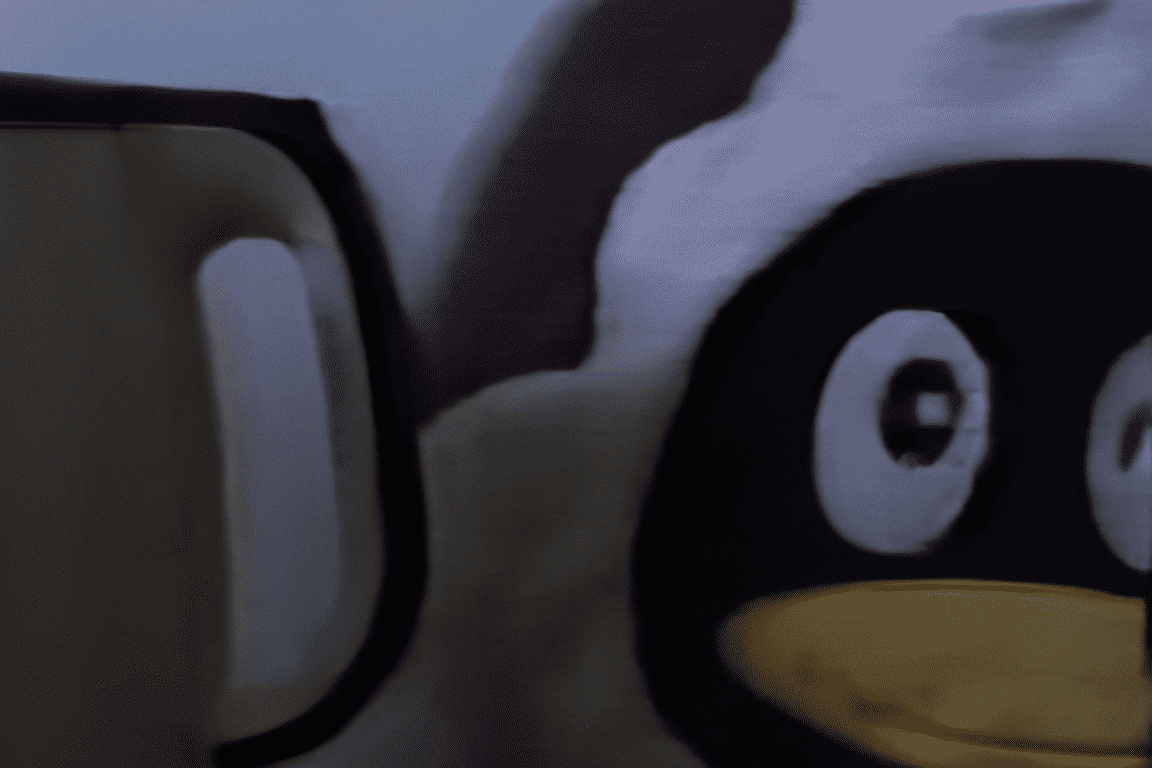}\hfill
			\includegraphics[width=\cimwid\linewidth,trim={0 0 0 0},clip]{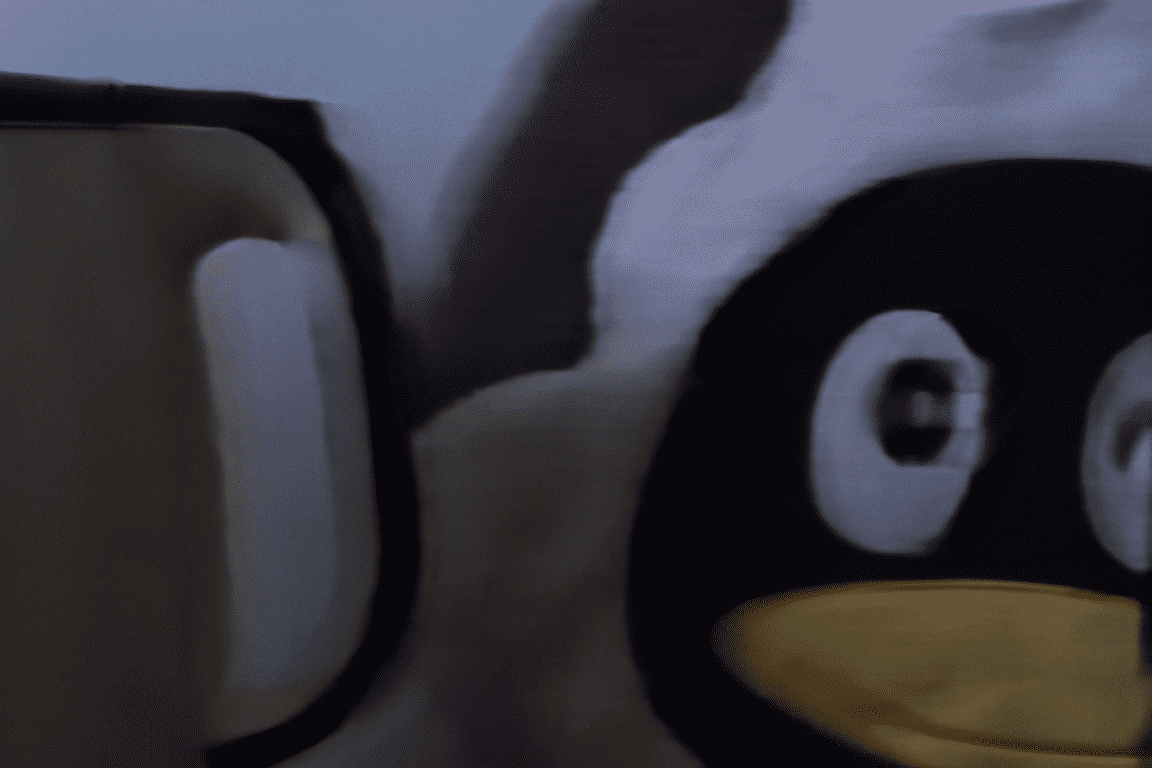}\hfill
    \end{minipage}
     \\
      \vspace{.2mm}
\begin{minipage}[t]{\linewidth}
    		\centering
      \begin{tikzpicture}[inner sep=0]
            \node[
            label= {[label distance=-0.13cm,text depth=2.3ex,rotate=90,align=center] left:\textcolor{black}{\footnotesize {(e)}}}]{\includegraphics[width=\ccimwid\linewidth,trim={0 0 0 0},clip,cframe=black .005mm]{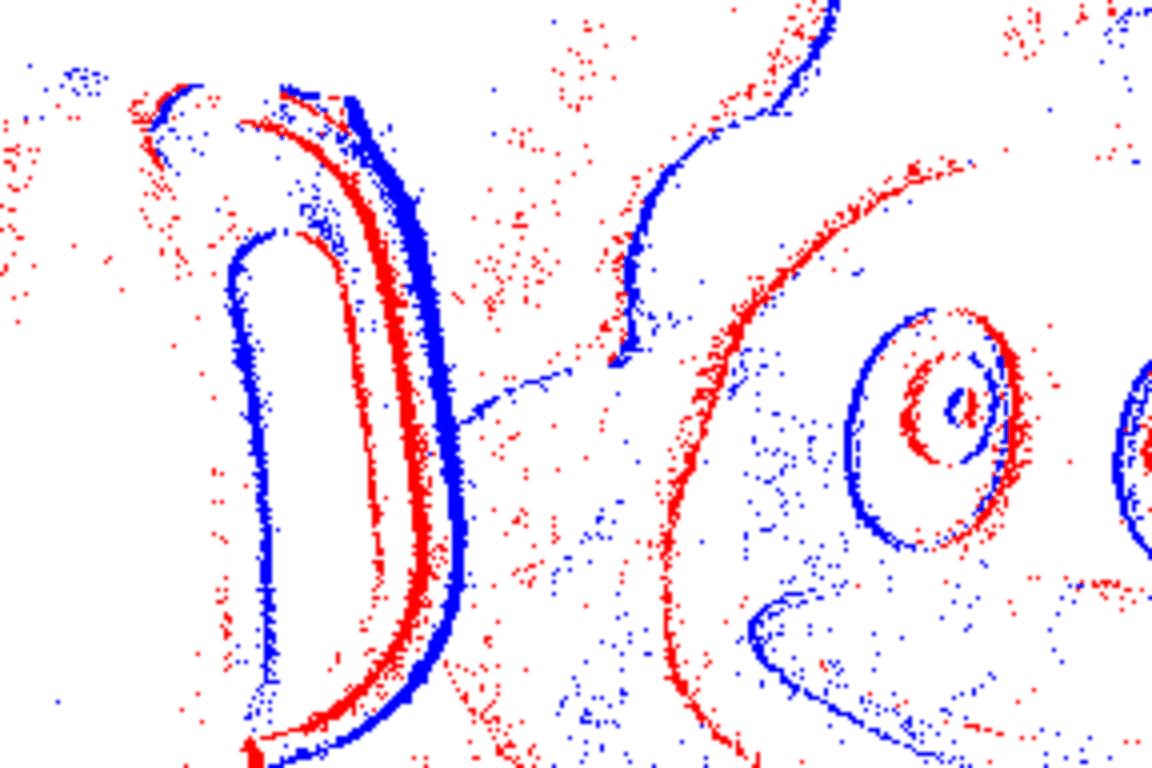}};
            \end{tikzpicture}\hfill
			\includegraphics[width=\ccimwid\linewidth,trim={0 0 0 0},clip,cframe=black .005mm]{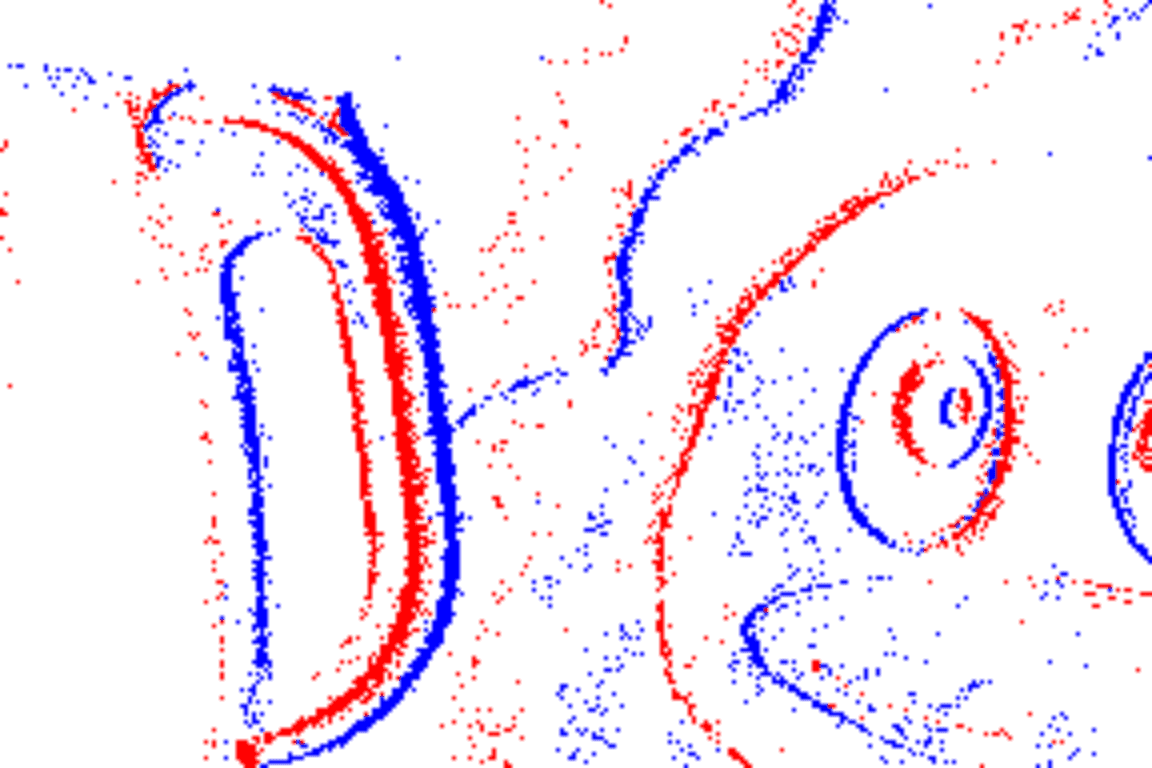}\hfill
			\includegraphics[width=\ccimwid\linewidth,trim={0 0 0 0},clip,cframe=black .005mm]{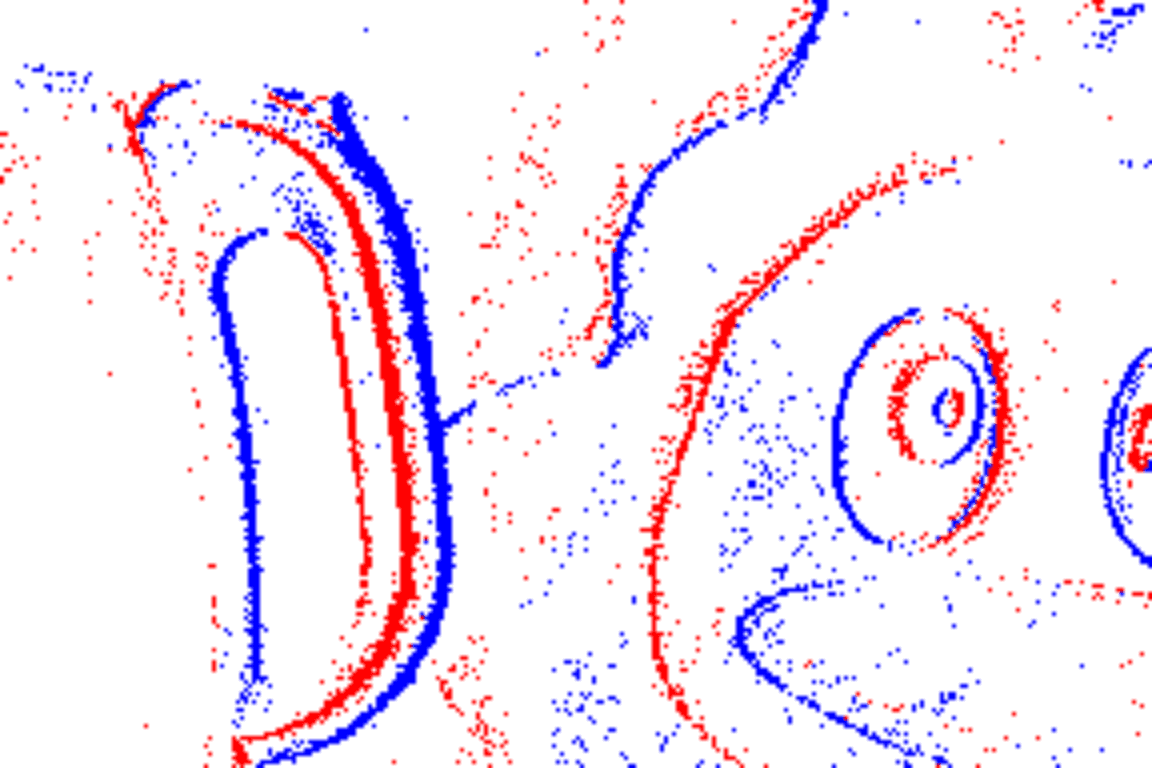}\hfill
			\includegraphics[width=\ccimwid\linewidth,trim={0 0 0 0},clip,cframe=black .005mm]{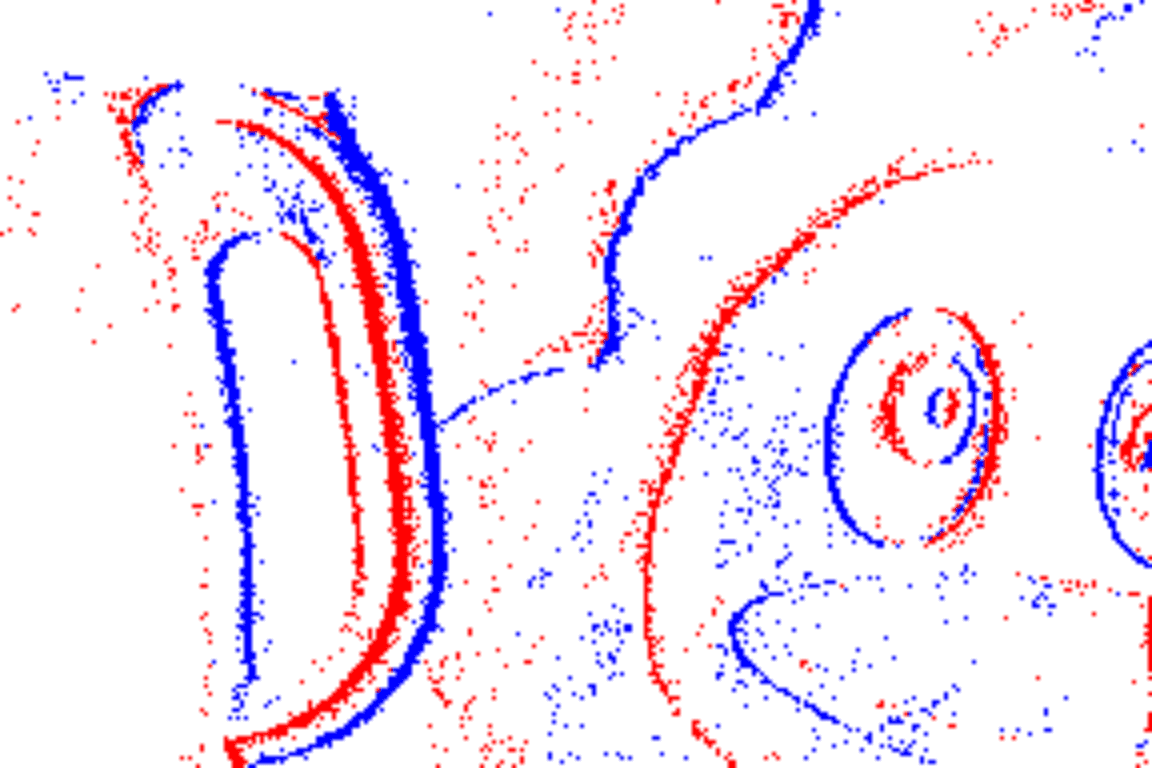}\hfill
			\includegraphics[width=\ccimwid\linewidth,trim={0 0 0 0},clip,cframe=black .005mm]{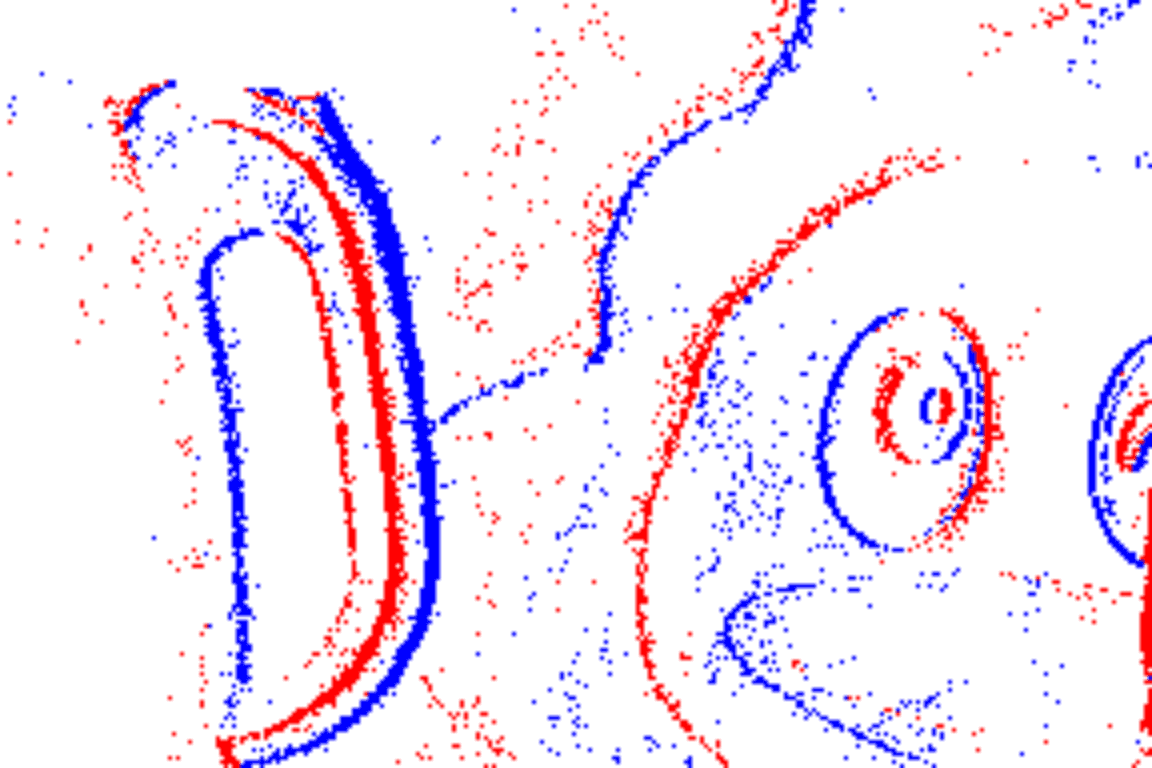}\hfill
			\includegraphics[width=\ccimwid\linewidth,trim={0 0 0 0},clip,cframe=black .005mm]{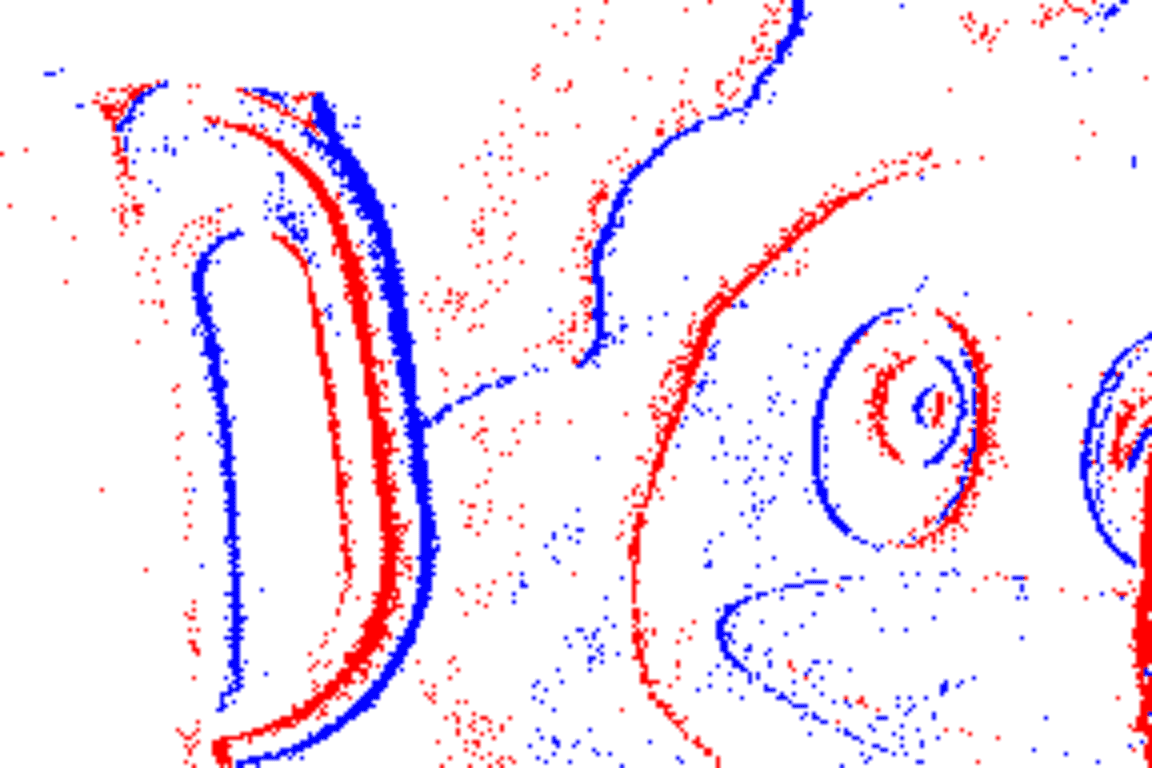}\hfill
			\includegraphics[width=\ccimwid\linewidth,trim={0 0 0 0},clip,cframe=black .005mm]{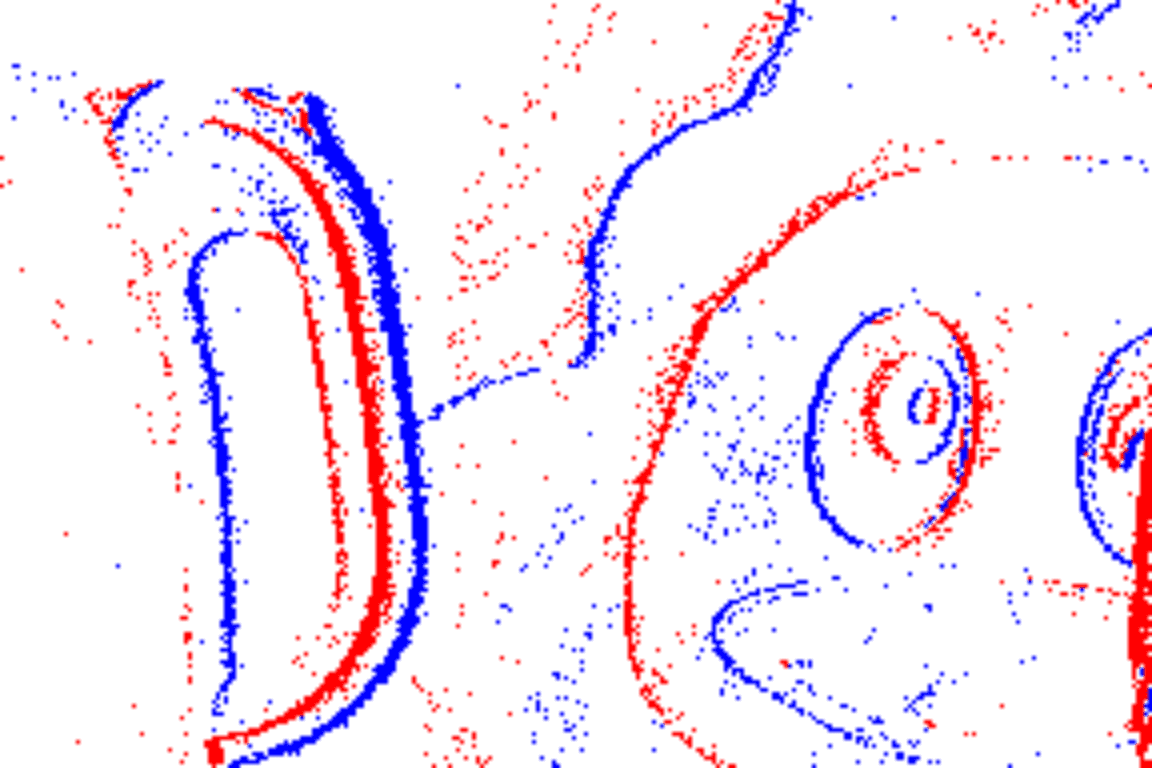}\hfill
    \end{minipage}
    \\
     \vspace{.4mm}
    \begin{minipage}[t]{\linewidth}
    		\centering
            \begin{tikzpicture}[inner sep=0]
            \node[
            label= {[label distance=0.05cm,text depth=-1ex,align=left] below:\textcolor{black}{\footnotesize {t = 1/8}}},
            label= {[label distance=-0.18cm,text depth=2.3ex,rotate=90,align=center] left:\textcolor{black}{\footnotesize {(f)}}}]{\includegraphics[width=\ccimwid\linewidth,trim={0 0 0 0},clip,cframe=black .005mm]{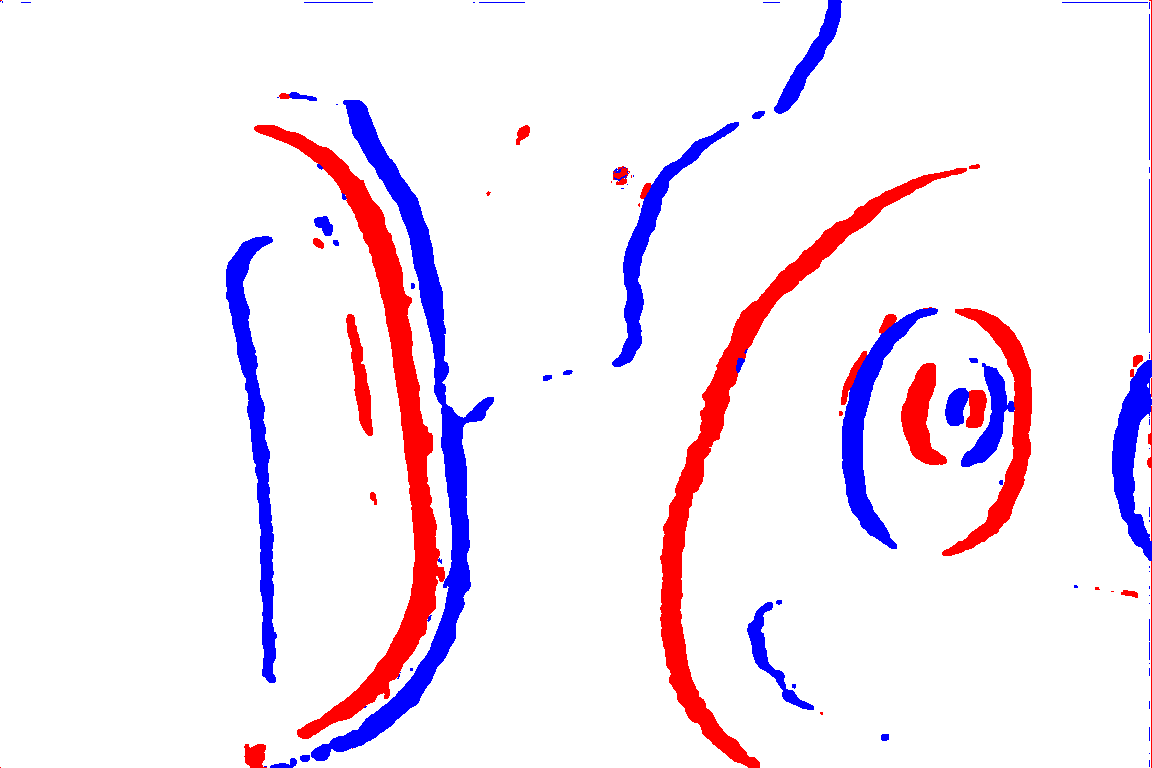}};
            \end{tikzpicture}
			\hfill
            \begin{tikzpicture}[inner sep=0]
            \node [label={[label distance=0.05cm,text depth=-1ex]below: \textcolor{black}{\footnotesize {t = 2/8}}}]  {
			\includegraphics[width=\ccimwid\linewidth,trim={0 0 0 0},clip,cframe=black .005mm]{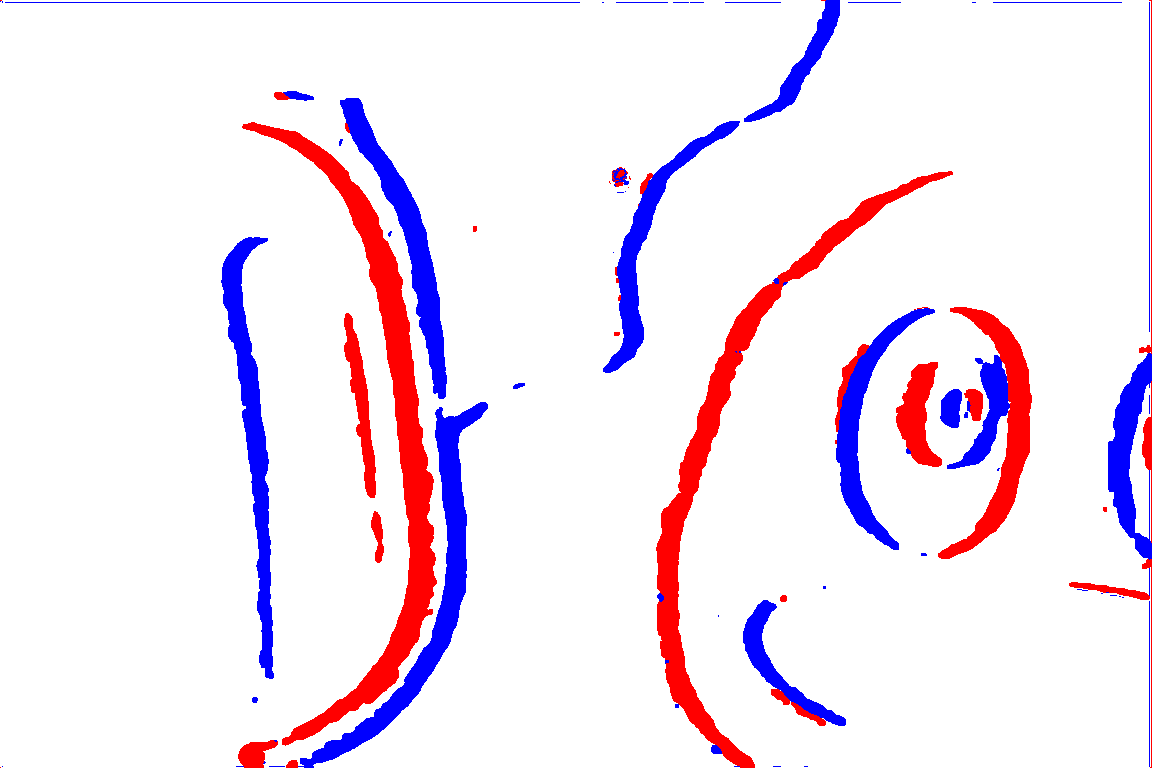}};\end{tikzpicture}
			\hfill
            \begin{tikzpicture}[inner sep=0]
            \node [label={[label distance=0.05cm,text depth=-1ex]below: \textcolor{black}{\footnotesize {t = 3/8}}}] {
			\includegraphics[width=\ccimwid\linewidth,trim={0 0 0 0},clip,cframe=black .005mm]{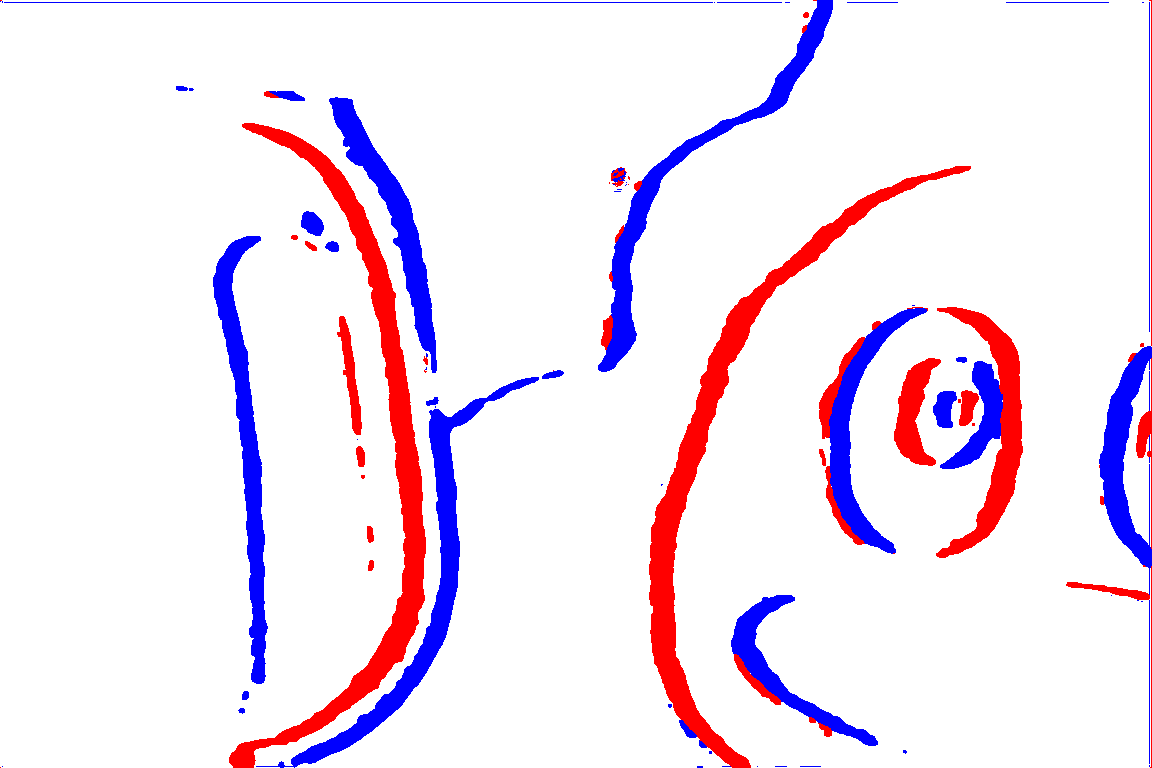}};\end{tikzpicture}
			\hfill
            \begin{tikzpicture}[inner sep=0]
            \node [label={[label distance=0.05cm,text depth=-1ex]below: \textcolor{black}{\footnotesize {t = 4/8}}}]  {
			\includegraphics[width=\ccimwid\linewidth,trim={0 0 0 0},clip,cframe=black .005mm]{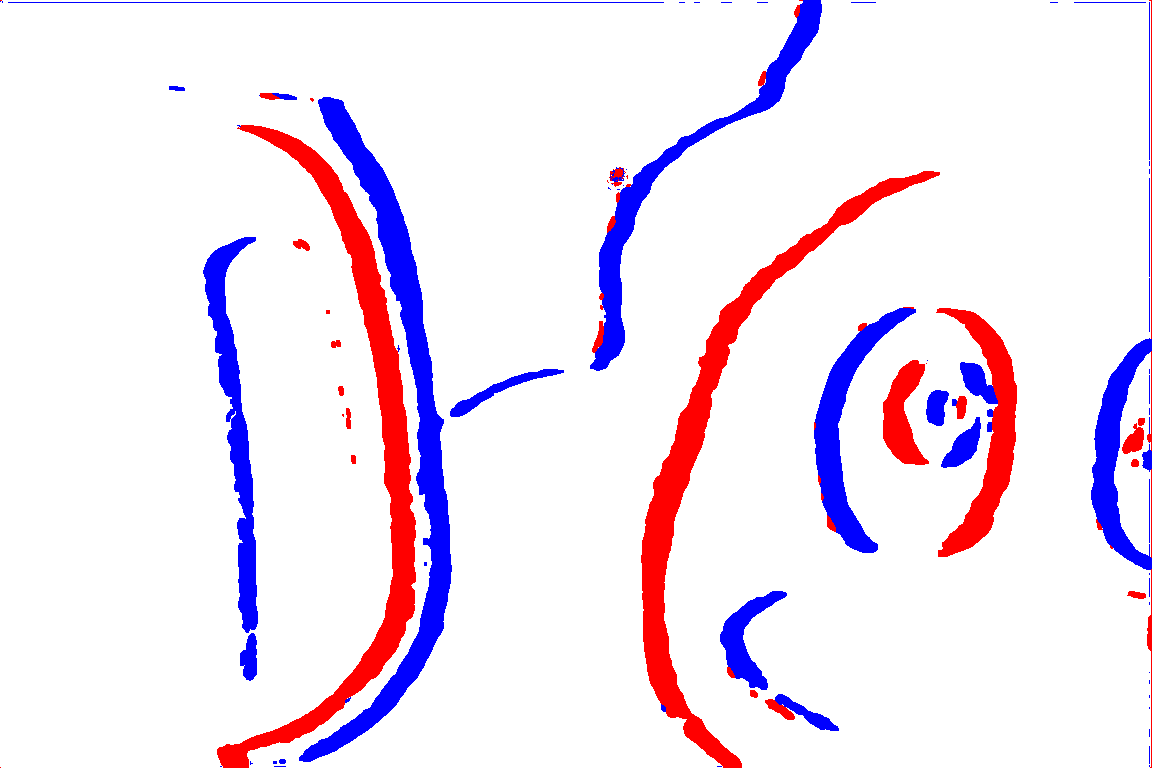}};\end{tikzpicture}
			\hfill
            \begin{tikzpicture}[inner sep=0]
            \node [label={[label distance=0.05cm,text depth=-1ex]below: \textcolor{black}{\footnotesize {t = 5/8}}}]  {
			\includegraphics[width=\ccimwid\linewidth,trim={0 0 0 0},clip,cframe=black .005mm]{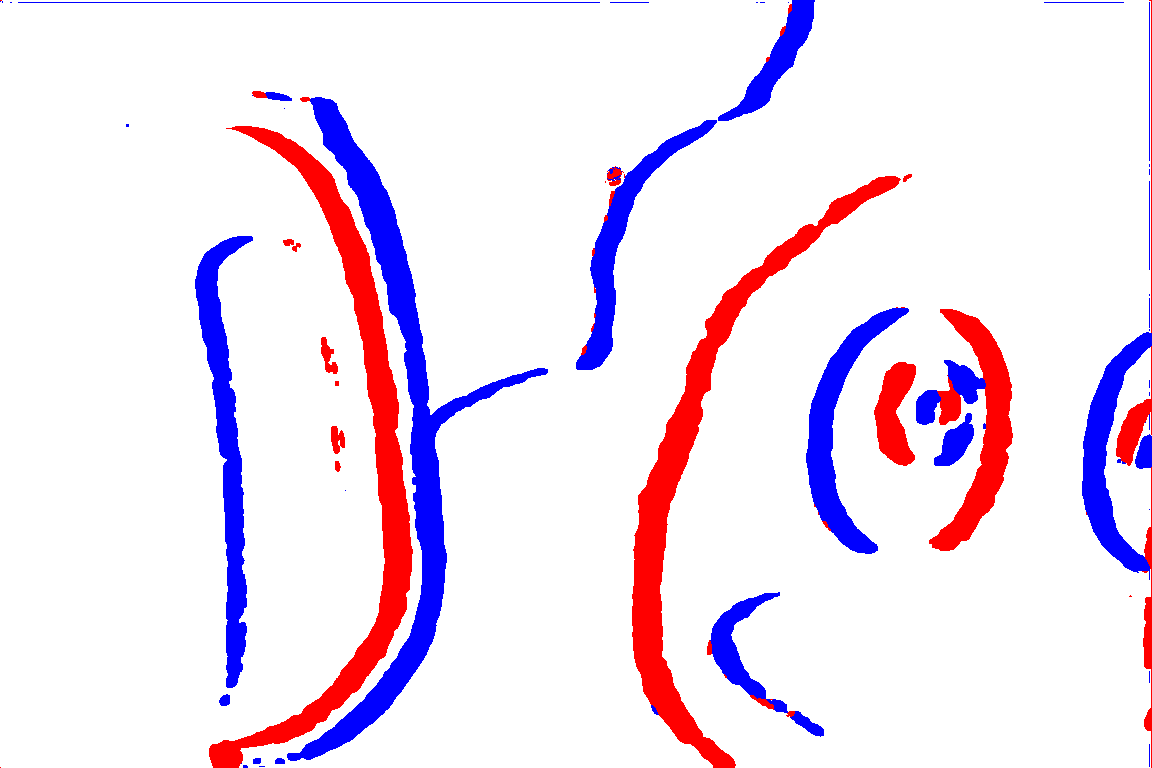}};\end{tikzpicture}
			\hfill
            \begin{tikzpicture}[inner sep=0]
            \node [label={[label distance=0.05cm,text depth=-1ex]below: \textcolor{black}{\footnotesize {t = 6/8}}}]  {
			\includegraphics[width=\ccimwid\linewidth,trim={0 0 0 0},clip,cframe=black .005mm]{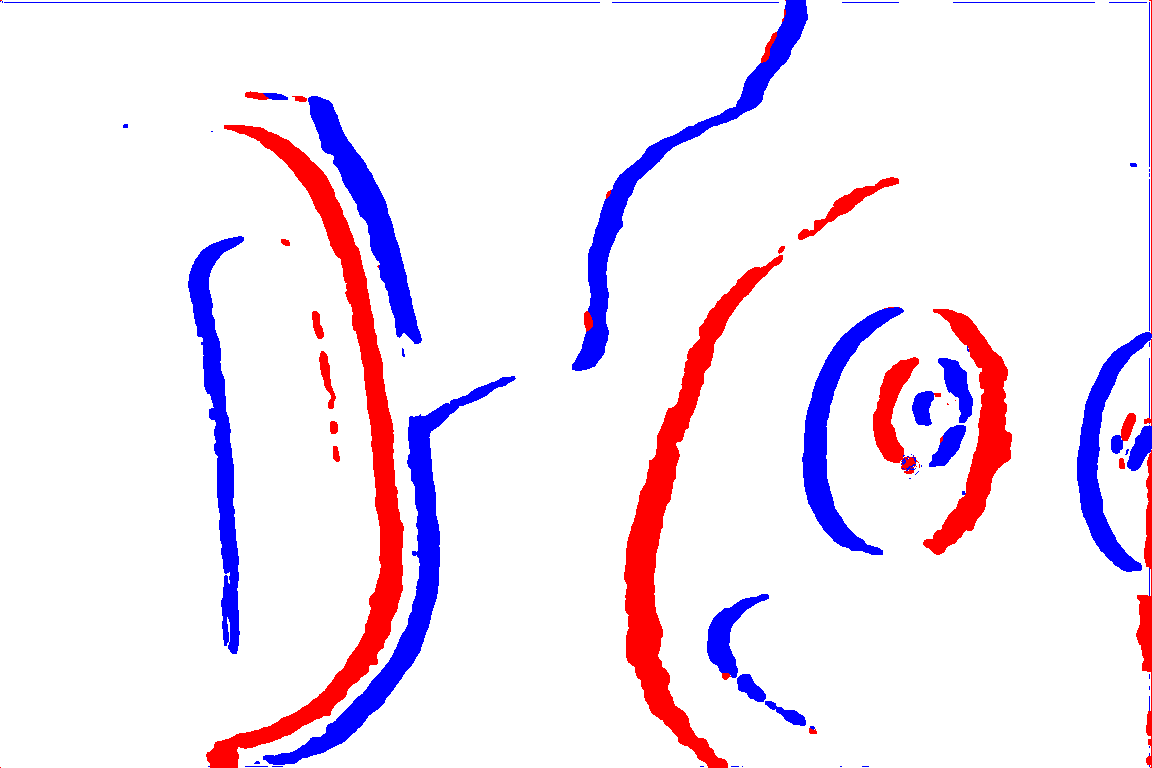}};\end{tikzpicture}
			\hfill
            \begin{tikzpicture}[inner sep=0]
            \node [label={[label distance=0.05cm,text depth=-1ex]below: \textcolor{black}{\footnotesize {t = 7/8}}}]  {
            \includegraphics[width=\ccimwid\linewidth,trim={0 0 0 0},clip,cframe=black .005mm]{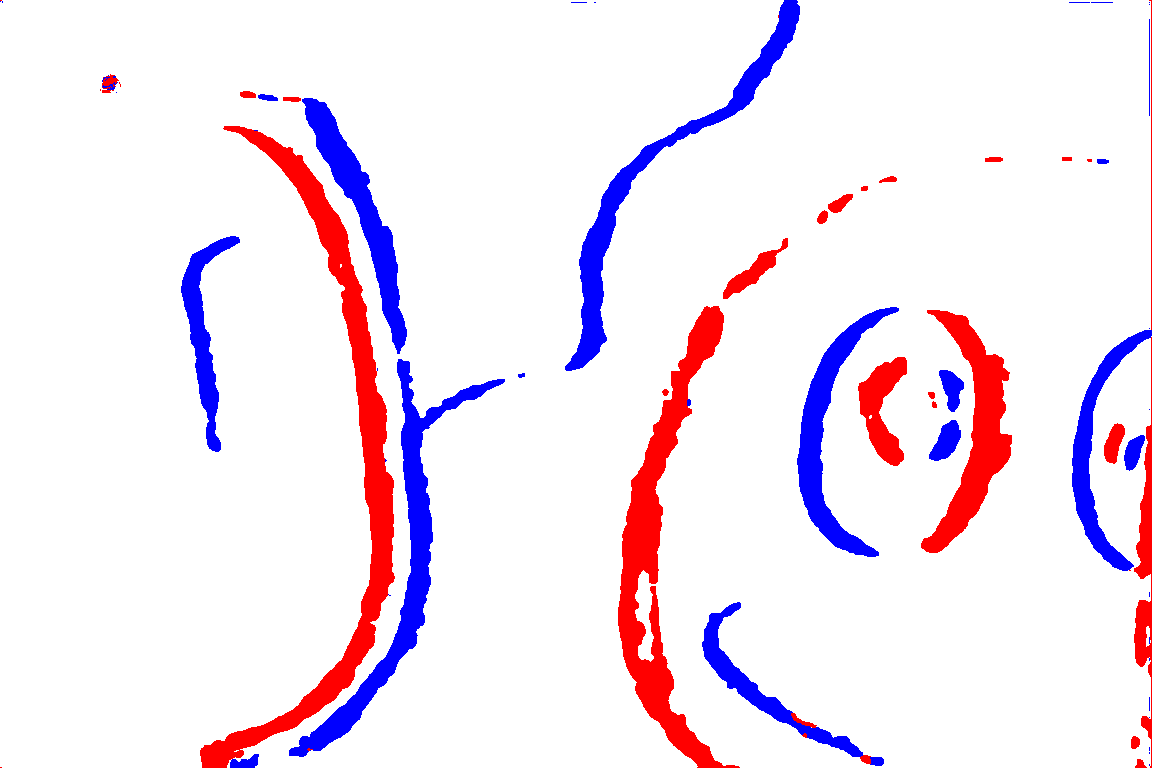}};\end{tikzpicture}
			\hfill
    \end{minipage}
	\caption{Illustration of our proposed CrossZoom Network (CZ-Net), which can restore a High-Resolution (HR) sharp latent sequence (d) and HR events (f) simultaneously from a single HR blurry image and corresponding Low-Resolution (LR) Events, and comparisons to the state-of-the-art methods, \eg, METR~\cite{zhang2021exposure} (c) and EZ+RED (b) by superimposing EventZoom (EZ)~\cite{duan2021eventzoom} for event super-resolving and RED~\cite{xu2021motion} for motion deblurring.
 }
	\label{motivation}
\end{figure}

\IEEEPARstart{N}{umerous} frame-event fusion-based approaches have been developed for diverse  applications~\cite{Guillermo2020PAMI,jiang2019mixed,zhang2022unifying,zhao2022transformer,wang2020joint}, capitalizing on the complementary spatiotemporal characteristics between conventional and event cameras. On the one hand, {\it E}vent-assisted methods {\it f}or {\it I}ntensity image enhancement, \ie, {\it the E4I approaches~\cite{jiang2019mixed,xu2021motion}},
commonly extract the inter/intra-frame information in terms of textures and motions as a complement benefiting from extremely high temporal resolution while disregarding their inherent limited spatial resolution~\cite{jing2021turning, wang2020event, han2021evintsr}. On the other hand, conventional intensity images are often considered as the guidance to denoise and super-resolve the noisy Low-Resolution (LR) event streams leading to {\it the I4E approaches}~\cite{zhao2022transformer,wang2020joint}. However, motion blurs in dynamic scenes often result in a significant loss in temporal resolution and are typically overlooked in I4E approaches.
The prevalent cross-modality inconsistency in the spatial resolution and occurrence of motion blur in degraded image frames (particularly in dynamic scenes with fast-moving objects) severely impair the performance~\cite{jing2021turning,wang2020event,zhao2022transformer,wang2020joint}. 

Despite the considerable advances in both E4I and I4E, limited exploration has been dedicated to simultaneously considering prevalent spatial resolution inconsistency and motion blur~\cite{Guillermo2020PAMI,zheng2023survey}. To boost the application of frame-event fusion-based approaches, a pressing need is emerging to simultaneously restore High-Resolution (HR) event streams and sharp images from a single HR blurry image and concurrent LR events, as shown in \cref{motivation}. Although one can achieve this by applying a two-stage cascaded approach that employs Event Super-Resolving (ESR) methods for LR events to assist existing Event-based Motion Deblurring (EMD) approaches, such a cascaded scheme often leads to sub-optimal performance since the deblurring or super-resolution errors will be propagated to the subsequent stage and inevitably degenerate the overall performance, as shown in \cref{motivation} (b).

Hence, we propose a unified architecture to accomplish the two tasks simultaneously, although its implementation is non-trivial. Specifically, the feasibility of this architecture is underpinned by the complementarity between HR blurry images and LR events. HR blurry images still contain valuable HR texture information capable of guiding ESR tasks~\cite{gao2019dynamic}, while the LR events encompass intra-frame motion and texture information that mitigate motion ambiguity in MD tasks. Nonetheless, this scheme introduces coupled distortions that increase the complexity of the task, \ie, coupled distortions primarily consist of motion ambiguity and texture erasure within the blurry image, as well as noise and texture discontinuity in the LR events arising from spatial degradation, as shown in ~\cref{motivation} (e). These intertwined complexities significantly intensify the ill-posedness of the MD and ESR tasks within a unified framework. Thus, the crucial challenge in architecting this unified cross-task scheme becomes evident: {\it effectively mitigating coupled distortions, leveraging the complementary advantages of blurry images and LR events, and achieving effective MD and ESR cohesively within this unified cross-task structure.}

To address these challenges, we propose a novel unified paradigm named the {\it C}ross{\it Z}oom {\it Net}work (CZ-Net) to achieve mutual enhancement in both spatial and temporal domains of LR events and blurry images by leveraging the complementary merits of multi-modal inputs, as shown in \cref{motivation}. Specifically, the Multi-scale Blur-Event Fusion (MBEF) strategy is introduced to fully exploit the scale-variant properties at different scales, guiding the super-resolving of LR events and incorporating the intra-frame information to alleviate the motion ambiguity in HR blurry image. To further harness the inherent complementarity between HR blurry images and LR events, we devise a Cross-Interaction Prediction (CIP) architecture to elevate HR intermediate features of MD and ESR branches by leveraging their respective spatiotemporal characteristics. Additionally, a blur-aware Attention-based Adaptive Enhancement (AAE) method is employed to effectively mitigate the coupled distortions and noise present in LR events and HR blurry images. 

Overall, the contributions of this paper are three-fold:

\begin{itemize}\setlength{\itemsep}{0.01pt}
    \item We propose a unified architecture, \ie, CZ-Net, to solve MD and ESR simultaneously. To the best of our knowledge, it is the first attempt to realize the sharp image sequence reconstruction and HR event stream generation by fusing a single HR blurry image and concurrent LR events. 
    \item We propose a cross-modality fusion module, \ie, MBEF, to achieve cross-enhancement by leveraging the scale-variant properties and effectively fusing cross-modality information. Meanwhile, blur-aware AAE and CIP modules are also presented to alleviate the distortions embedded in the LR events and reinforce the final results through the prior blur-event complementary information.
    \item We build a new Cross-Resolution Deblurring and Resolving (CRDR) dataset composed of paired HR {\it sharp-blurry} images and {\it HR-LR} event streams under various indoor and outdoor scenes. Extensive qualitative and quantitative comparisons demonstrate the effectiveness and robustness of our proposed CZ-Net. Codes and datasets are released at \href{https://bestrivenzc.github.io/CZ-Net/}{https://bestrivenzc.github.io/CZ-Net/} to facilitate future research. 
\end{itemize}

\section{Related Work}
\subsection{Frame-based Motion Deblurring} 
The frame-based Motion Deblurring (MD) task aims to recover latent sharp images from blurry input, which is ill-posed due to the missing intra-frame motions and textures. Traditional deblurring methods often formulate the task as an inverse problem, which models a blurry image as the result of convolution with the blur kernels and can be roughly divided into blind~\cite{fergus2006removing,michaeli2014blind} and non-blind~\cite{cho2011handling,schmidt2013discriminative,szeliski2022computer,schmidt2013discriminative,xu2014inverse} architectures. Some early non-blind approaches adopt classical image deconvolution algorithms such as Wiener deconvolution, Lucy-Richardson with or without Tikhonov regularization, to restore sharp images~\cite{szeliski2022computer,schmidt2013discriminative,xu2014inverse}. On the other hand, blind image deblurring methods~\cite{chan1998total,dong2017blind,chen2019blind} aim to recover both the sharp image and the blur kernel simultaneously. In addition, various priors, including Total Variation (TV)~\cite{chan1998total}, outlier image signals~\cite{dong2017blind}, and local maximum gradient~\cite{chen2019blind} are designed to relieve the burden of ill-posedness. Generalization properties and computational efficiency usually limit the performance of these methods.

Recently, deep neural network approaches have directly learned to map blurry images to sharp latent restorations from paired datasets, achieving dominant performance~\cite{gao2019dynamic,cho2021rethinking,nah2017deep,arjomand2017deep, Kupyn2018CVPR, Zamir2021MPRNet}. Techniques, including coarse-to-fine strategies~\cite{nah2017deep,cho2021rethinking} and the imposing artificial~\cite{arjomand2017deep} or data-driven~\cite{Kupyn2018CVPR} priors, have been investigated to improve the performance of latent sharp image reconstruction. To optimally balance spatial details and high-level contextualized information, MPR-Net~\cite{Zamir2021MPRNet} designs a multistage architecture that progressively learns restoration functions. Despite these advances, all methods mentioned above format the data-driven or optimize-based frameworks on the blurry inputs with the missing intra-frame motion and texture, leading to marginal performance and limited model generalization.  

\begin{figure*}[t]
\centering
\includegraphics[scale=0.61]{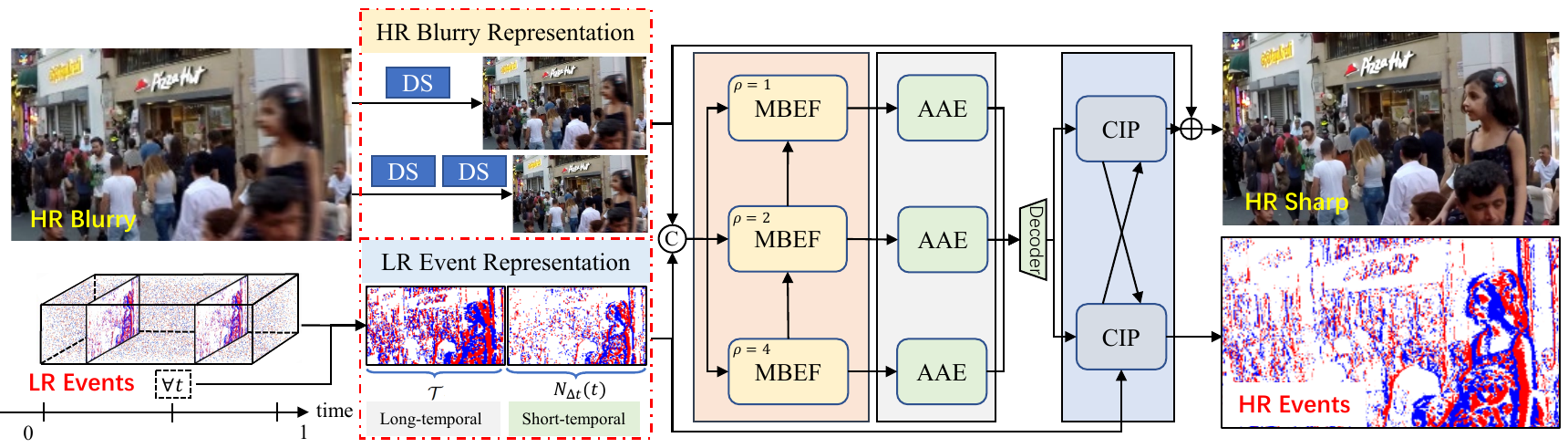}
\caption{Architecture of the proposed CZ-Net, which is composed of a Multi-scale Blur-Event Fusion (MBEF) module, an Attention-based Adaptive Enhancement (AAE) module, and a Cross-Interaction Prediction (CIP) module.}
\label{framework}
\end{figure*}

\subsection{Event-based Image Deblurring}
Event sensors are used for image deblurring due to their high temporal resolution~\cite{pan2019bringing,pan2020high}. The first work on Event-based Motion Deblurring (EMD)~\cite{pan2019bringing} establishes the relationships between blurry images, events, and latent frames using an Event-Based Double Integral (EDI) model, which often suffers from heavily accumulated noise, leading to inaccurate details. To mitigate defects in~\cite{pan2019bringing}, the multiple EDI (mEDI) is proposed to get a less noisy estimation from multiple images and events~\cite{pan2020high}. Learning-based approaches for EMD significantly reduce the disturbance of event noise~\cite{xu2021motion,zhang2022unifying,sun2022event}. Specifically, U-Net based on dual encoders is designed in~\cite{haoyu2020learning} to perform image deblurring followed by high frame rate video generation. To alleviate performance degradation caused by data inconsistency, Xu \etal~\cite{xu2021motion} design a semi-supervised framework using real-world events. Recently, the two-stage image restoration network EF-Net~\cite{sun2022event} combines a cross-modal attention module and symmetric cumulative event representation to fuse event and image effectively. However, the aforementioned methods only focus on restoring sharp latent images from motion-blurred images with the same spatial resolution as the events. They cannot deal with scenarios when events and blurry images have different scales.

\subsection{Event Super-Resolving} 
The event stream acquired by detecting intensity changes is essentially a kind of spatio-temporal data. Accurate and high spatiotemporal distribution needs to be estimated for LR Event Super-Resolving (ESR). A few works~\cite{li2019super,wang2020joint,duan2021eventzoom,weng2022boosting} are developed to achieve ESR. A multi-stage scheme is proposed in~\cite{li2019super}, sampling the events of each pixel by simulating a Poisson point process according to the specified event number and modeling a spatiotemporal filter to generate the temporal rate function. The beneficial information in HR image signals is used to guide the process of the events upsampling in an optimization framework termed GEF~\cite{wang2020joint}, whereas the performance of GEF is intricately linked to both the sharpness of the image and the accuracy of optical flow prediction. Duan \etal~\cite{duan2021eventzoom} develop a multi-task framework in denoising and super-resolving neuromorphic events, realizing the denoising and the upsampling mapping in a noise-to-noise manner. More recently, Weng \etal~\cite{weng2022boosting} propose a recurrent neural network for reliably aggregating the spatio-temporal clues of event stream and a temporal propagation for incorporating neighboring and long-range event-aware contexts. However, they restore the HR event streams with limited temporal resolution and insufficiently accurate polarity owing to the fixed channel of the representations of the events.

\section{Method}

\subsection{Problem Formulation}
We first revisit the fundamental models for Motion Deblurring (MD) and Event Super-Reoslving (ESR), then formulate the task of the unified Event-based motion Deblurring and Super-Resolving (uEDSR), for which we propose the CrossZoom Network (CZ-Net) in the subsequent subsection. 

\noindent\textbf{Motion Deblurring.} The blurry image $\mathbf{B}$ can be formulated as the average of the latent sharp images $\mathbf{I}$ within the exposure interval $\mathcal{T}$, \ie, $\kappa: \mathcal{I}\times \mathcal{T}\to \mathcal{I}$ with $\mathcal{I}$ denoting the image subspace, leading to a significant loss in temporal resolution. 
\begin{equation}\label{ED}
    \mathbf{B}=\kappa(\mathbf{I}) \triangleq \frac{1}{T}\int_{t\in{\mathcal{T}}}\mathbf{I}(t)dt,
\end{equation}
with $T$ representing the duration of $\mathcal{T}$. Motion Deblurring (MD) aims at zooming in the temporal domain, \ie, recovering sharp latent images $\{\mathbf{I}(t)\}_{t\in{\mathcal{T}}}$ from the blurred image $\mathbf{B}$, 
\begin{equation}\label{EMD-Net}
    \{\mathbf{I}(t)\}_{t\in{\mathcal{T}}}=\operatorname{Deblur}(t;\mathbf{B},  \mathcal{E}_{\mathcal{T}}),
\end{equation}
where $\operatorname{Deblur}: \mathcal{I}\to \mathcal{I}\times\mathcal{T}$ denotes an MD operator conditioned upon the auxiliary event streams $\mathcal{E}_{\mathcal{T}}\triangleq \{(\mathbf{x}_i,t_i,p_i)\}_{t_i\in \mathcal{T}}$, with $\mathbf{x}_i$ the pixel location, $t_i$ the time of the $i$-th event occurs, and $p_i\in\{-1,+1\}$ the polarity, representing brightness change directions~\cite{Guillermo2020PAMI}. Note that existing approaches~\cite{xu2021motion,zhang2022unifying} are only feasible when events $\mathcal{E}_{\mathcal{T}}$ share the same scale as the blurry image $\mathbf{B}$.





\noindent\textbf{Event Super-Resolving.} Ideally, the degradation of events from High-Resolution (HR) to Low-Resolution (LR) can be represented as a spatial downsampling procedure,
\begin{equation}\label{ESR-1}
    \mathcal{E}_\mathcal{T}^{\rho} = \eta(\mathcal{E}_\mathcal{T}, \rho) \triangleq \{(\mathbf{x}_i,t_i,p_i)| (\mathbf{x}_i / \rho,t_i,p_i) \in \mathcal{E}_\mathcal{T}\},
\end{equation}
with $\rho \in \{1,2,3,...\}$ denoting the rescale factor and $\eta$ indicating downsampling operator. Addressing the issue of ESR entails the prediction of the HR event stream $\mathcal{E}_\mathcal{T}$ from its LR counterpart $\mathcal{E}_\mathcal{T}^{\rho}$, which accomplishes zooming in the spatial domain for LR events, \ie,
\begin{equation}\label{ESR}
    \mathcal{E}_\mathcal{T}=\operatorname{SuperR}(\mathcal{E}_\mathcal{T}^{\rho}, \mathbf{I}),
\end{equation}
where $\operatorname{SuperR(\cdot)}$ indicates an ESR operator with the inclusion of sharp frames $\mathbf{I}$ as the auxiliary information. However, existing frame-assisted ESR approaches degrade severely when $\mathbf{I}$ deteriorates in the temporal domain as~\cref{EMD-Net} due to the vanishing of HR sharp details~\cite{wang2020joint}. Moreover, limited temporal resolution and insufficiently accurate polarity of super-resolved events still prevent the applicability of existing ESR methods~\cite{duan2021eventzoom,weng2022boosting}.    


\begin{figure}[t]
\centering
\includegraphics[scale=0.88]{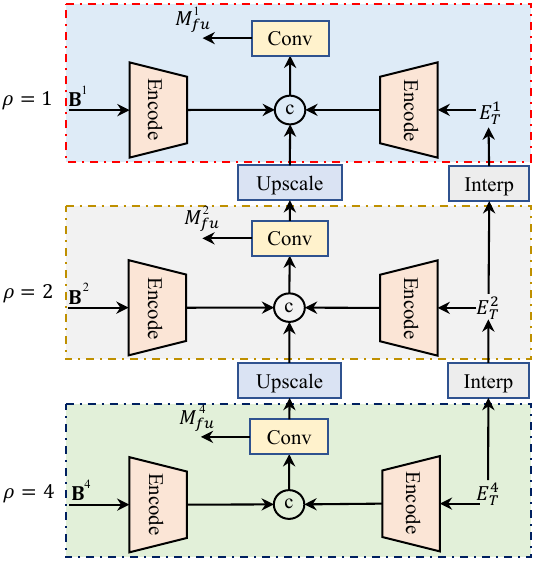}
\caption{Details of Multi-scale Blur-Event Fusion (MBEF) module, which integrates the blurry images and the event feature at multi-scales.}
\label{ITE_MBEF}
\end{figure}

\noindent\textbf{Unified Task.} The unified task is to accomplish both MD and ESR simultaneously, \ie, 
\begin{equation}\label{EDSR}
    \{\mathbf{I}(t)\}_{t\in{\mathcal{T}}},\mathcal{E}_{\mathcal{T}}=\operatorname{uEDSR}(t;\mathbf{B},\mathcal{E}_\mathcal{T}^{\rho}),\quad \forall{t}\in{\mathcal{T}}.
\end{equation}
where $\operatorname{uEDSR}$ denotes the {\it u}nified {\it E}vent-based motion {\it D}eblurring and {\it S}uper-{\it R}esolving (uEDSR) operator. Note that replacing $\mathbf{B}$ by $\mathbf{I}$ will degenerate $\operatorname{uEDSR}$ into the single ESR task as \cref{ESR} and $\rho=1$ corresponds to the single MD task as \cref{EMD-Net}. However, the coexistence of degenerations for the image in the temporal domain and events in the spatial domain makes uEDSR more challenging than MD or ESR.  
Although it is feasible to accomplish uEDSR through a two-stage process by cascading ESR and MD without the aid of cross modalities, it often leads to suboptimal performance due to error propagation, as shown in \cref{motivation} (b). 


\subsection{CrossZoom Network}

We propose a novel CrossZoom Network (CZ-Net) to accomplish uEDSR by exploring the mutual compensations from the degraded cross-modalities, \ie, the blurry image $\mathbf{B}$ and the LR events $\mathcal{E}_{\mathcal{T}}^{\rho}$. \cref{framework} illustrates the framework of the proposed CZ-Net, which is a deep network consisting of a Multi-scale Blur-Event Fusion (MBEF) module, an Attention-based Adaptive Enhancement (AAE) module, and a Cross-Interaction Prediction (CIP) module.

\subsubsection{Spatiotemporal Representation}
To fully exploit the respective advantages of HR images $\mathbf{B}$ in the spatial domain and LR event streams $\mathcal{E}_{\mathcal{T}}^{\rho}$ in the temporal domain, we sample both $\mathbf{B}$ and $\mathcal{E}_{\mathcal{T}}^{\rho}$ at multiple scales. 

\noindent\textbf{Blurry Image Representation.} 
Due to the scale-variant representation, where blurred edges become sharper after scaling down~\cite{gao2019dynamic}, we extract multi-scale sharpness information inherent in the blurry inputs $\mathbf{B}^1$ by DownSampling (DS) them to $\mathbf{B}^2$ and $\mathbf{B}^4$ with smaller scales. Here, the subscript numbers, $2$ and $4$, represent the scale reduction factors, as shown in ~\cref{framework}.

\noindent\textbf{Event Representation.} We encodes LR event streams $\mathcal{E}^\rho_\mathcal{T}$ into time-dependent representation frames $E^{\rho}_{\Delta{t}}(t)$ and entire-time event representation frames $E^{\rho}_{\mathcal{T}}$, denoted as:
\begin{equation}\label{ER}
\setlength{\abovedisplayskip}{5pt}
\setlength{\belowdisplayskip}{5pt}
\begin{split}
    E^{\rho}_{\Delta{t}}(t)&=\operatorname{ER}(\mathcal{E}^{\rho}_{\mathcal{N}_{\Delta{t}}(t)}) \\
    E^{\rho}_{T}&=\operatorname{ER}(\mathcal{E}^{\rho}_{\mathcal{T}}),
\end{split}
\end{equation}
where $\operatorname{ER}(\cdot)$ denotes the {\it E}vent {\it R}epresentation operator, while $\mathcal{N}_{\Delta{t}}(t)\triangleq \{t^\prime\mid\left|t-t^\prime\right|\le{\frac{\Delta{t}}{2}},\forall t^\prime\in{\mathcal{T}}\}$ defines the interval centered around the timestamps $t$. The time-dependent representation, $E^{\rho}_{\Delta{t}}(t)$, implicitly encodes timestamp information, which is crucial to guide the recovery of sharpness at specific given timestamps $t$. On the contrary, the whole-time representation, $E^{\rho}_{\mathcal{T}}$ incorporates comprehensive motion information spanning the entire exposure time interval $\mathcal{T}$ and effectively mitigates motion ambiguity in the MD task. The final event representation frame $E^{\rho}$ is obtained by concatenating the $E^{\rho}_{\Delta{t}}(t)$ and $E^{\rho}_{\mathcal{T}}$, and we employ the Voxel Grid method~\cite{zihao2018unsupervised} as $\operatorname{ER}$ to quantize information into eight bins.

\subsubsection{Multi-scale Blur-Event Fusion}
Another challenge for CZ-Net is the effective utilization of complementarities between degraded inputs $\mathbf{B}$ (suffering from temporal collapse) and $\mathcal{E}^{\rho}_{\mathcal{T}}$ (undergoing spatial collapse) which have a different spatial resolution. Hence, the MBEF module is designed to perform cross-enhancement by fusing $B^{R}$, which holds spatial multi-scale sharpness information, with $E^{R}$, containing temporal multi-scale intra-frame motion across various scales. Here, $R\triangleq \{1,2,4\}$ represents the set of scale reduction factors.


Specifically, as depicted in~\cref{ITE_MBEF}, we initiate the process by obtaining $E^{R}$ at various scales by applying two bilinear interpolation operators. Subsequently, we extract multi-scale blurry and event features, denoted as $M^{R}_{b}$ and $M^{R}_{e}$, respectively, by utilizing dedicated blur- and event-encoding networks. Then we derive the blur-event fusion feature $M^{4}_{fu}$ along with its associated upscaled feature $U^{2\times}$, denoted as: 
\begin{equation}\label{MBEF-1}
\setlength{\abovedisplayskip}{5pt}
\setlength{\belowdisplayskip}{5pt}
\begin{split}
    M_{fu}^{4}&=\operatorname{Conv}(M_{b}^{4}\oplus M_{e}^{4}) \\
    U^{2}&=o(\operatorname{Upscale}(M_{fu}^{4})), \\
\end{split}
\end{equation}
where the symbol $\oplus$ denotes the concatenation operation; the operator $\operatorname{Conv}(\cdot)$ corresponds to double dilated {\it Conv}olutional layers followed by the activation function $\operatorname{Relu}$; the $\operatorname{Upscale}(\cdot)$ means the use of a transposed layer for feature upscaling with trainable parameters; and the mapping function $o(\cdot)$ is required for generating an intermediate upscaled feature map, which is achieved through a single convolutional layer. We extend this procedure to encompass blur-event fusion features $M^{2}_{fu}$ with scale reduction factor of $2$, with the omitting of $1$ for simplicity, which can be formulated as
\begin{equation}\label{MBEF-2}
\setlength{\abovedisplayskip}{5pt}
\setlength{\belowdisplayskip}{5pt}
\begin{split}
    M_{fu}^{2}&=\operatorname{Conv}(M_{b}^{2}\oplus M_{e}^{2} \oplus U^{2}) \\
    U^{1}&=o(\operatorname{Upscale}(M_{fu}^{2})). \\
\end{split}
\end{equation}
The concatenation terms incorporate information from the previous scale $U^{2}$, enhancing the fusion feature $M_{fu}^{2}$. By employing the MBEF module, we acquire fusion features $M^{R}_{fu}$ that effectively leverage the complementarities between $\mathbf{B}$ and $\mathcal{E}_{\mathcal{T}}^{\rho}$, facilitating simultaneous cross-zooming in the temporal domain for blurry image $\mathbf{B}$ and in the spatial domain for LR events $\mathcal{E}_{\mathcal{T}}^{\rho}$.

\begin{figure}
\centering
\includegraphics[scale=0.76]{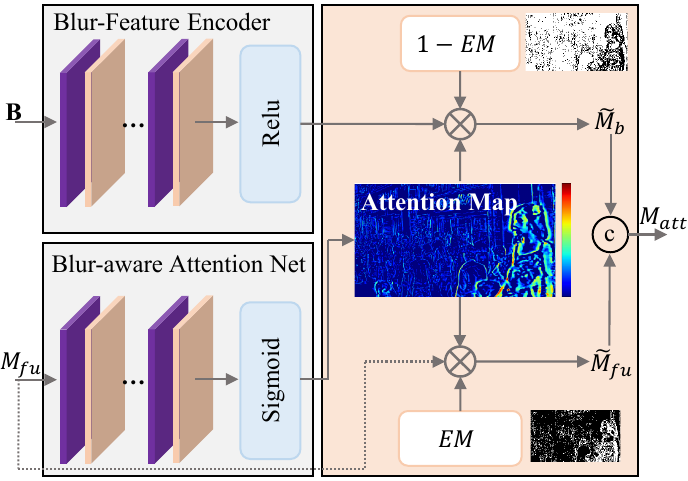}
\caption{Architecture of the Attention-based Adaptive Enhancement (AAE) module, equipped with a blur-aware attention and Event Mask (EM) to alleviate the coupled distortions embedded in HR blurry image and LR events.}
\label{AFA}
\vspace{-.9em}
\end{figure}

\subsubsection{Attention-based Adaptive Enhancement}
The fusion-based features $M_{fu}$ estimated by the MBEF operator still suffer from the imperfections of LR events for two primary reasons: i) Values in $M_{fu}$ that correspond to the sparse regions in LR events are inferred from surrounding isolated values, achieved by incrementally expanding receptive fields; ii) Portions of representations within $M_{fu}$ originate from noise inherent in LR events. Both situations result in low-confidence, unreliable features, which pose challenges for subsequent tasks such as MD and ESR. To this end, we introduce the Attention-based Adaptive Enhancement (AAE) module, designed to incorporate the $M_{fu}$ and $\mathbf{B}$ adaptively by using a blur-aware attention map and an event mask. This process assigns less attention to invalid representations in $M_{fu}$ while enhancing the confidence of the features derived from $\mathbf{B}$ that provide valuable clues, including multi-scale sharpness information. As illustrated in \cref{AFA}, the proposed AAE module consists of two main steps: Blur-Aware Attention Generation and Adaptive Fusion. These steps take $M_{fu}$ and the blurry image $\mathbf{B}$ as inputs and produce an adaptive attention-based feature denoted as $M_{att}$. For brevity, we omit superscripts that indicate different scale factors.

\noindent\textbf{Blur-Aware Attention Generation.} Blur-aware attention map is used as a gating function to deliver more weight to blurry pixels that equal the magnitude of motion. Let $M_{fu}$ be an input tensor, $\operatorname{BA-Net}$ a Blur-aware Attention Network that learns the map from $M_{fu}$ to a significant vector $Y=\operatorname{BA-Net}(M_{fu})$, consisting of a double $\operatorname{Conv}$ operator in ~\cref{MBEF-1}. An attention map $A$ can be computed by
\begin{equation}
\setlength{\abovedisplayskip}{5pt}
\setlength{\belowdisplayskip}{5pt}
\begin{split}
    A = \sigma\left(Y\right)=\operatorname{Sigmoid}\left(\operatorname{BA-Net}\left(M_{fu}\right)\right).
\end{split}
\end{equation}
Akin to~\cite{qian2018attentive}, we take the difference between the blurry and sharp frames at timestamps $t$ and subsequently apply thresholding to generate the ground-truth blur mask. Furthermore, instead of applying the softmax, which is the sum-to-one constraint, we use a sigmoid activation function which only constrains attention importance values ranging from 0 to 1: $A=1/(1+\operatorname{exp}(-Y))$. Then, we supervise the $\operatorname{BA-Net}$ by the pixel-wise $l_1$ norm loss as
\begin{equation}
\setlength{\abovedisplayskip}{5pt}
\setlength{\belowdisplayskip}{5pt}
\begin{split}
    \mathcal{L}_{att}=\Vert{\hat{A}-A}\Vert{_1},
\end{split}
\end{equation}
where $\hat{A}$ is the difference map between latent sharp and blurry images. Consequently, the attention map $A$ contains blurry information that denotes the confidence of the events emitted.

\noindent\textbf{Adaptive Enhancement.}
The blur-aware attention map $A$, in conjunction with the binary {\it E}vent {\it M}ask denoted as $EM$ that signifies the occurrence or absence of emitted events, is employed to assign lower attention values to areas with no motion blur yet events activated for suppressing the invalid feature in $M_{fu}$ caused by the imperfection of noise and sparsity. Mathematically, we can calculate the enhanced attention-based feature $\tilde{M}_{fu}$ as  
\begin{equation}
\setlength{\abovedisplayskip}{5pt}
\setlength{\belowdisplayskip}{5pt}
\begin{split}
    \tilde{M}_{fu}=[W_{1}\otimes{M_{fu_{1}}},W_{1}\otimes{M_{fu_{2}}},...,W_{1}\otimes{M_{fu_{m}}}],
\end{split}
\end{equation}
where $W_1=A\otimes{EM}$ denotes the weighting map calculated from blur-aware attention map $A$ and event mask $EM$ with a value of 1 assigned to event pixels and 0.1 to non-event-pixels, $\otimes$ is an element-wise multiplication, and suffix $m$ represents the channel number of $M_{fu}$. Furthermore, for areas in $M_{fu}$ where events are absent, but motion blurs exist, the feature $M_b$ obtained by a blur-feature encoder, consisting of a double $\operatorname{Conv}$ operator in ~\cref{MBEF-1}, contains multi-scale sharpness information and is given more attention to enhancing the deblurring results. The enhanced feature $\tilde{M}_{b}$ can be represented as
\begin{equation}
\setlength{\abovedisplayskip}{5pt}
\setlength{\belowdisplayskip}{5pt}
\begin{split}
    \tilde{M}_{b}=[W_{2}\otimes{M_{b_{1}}},W_{2}\otimes{M_{b_{2}}},...,W_{2}\otimes{M_{b_{n}}}],
\end{split}
\end{equation}
where $W_2=A\otimes{(1-EM)}$ and the suffix $n$ means the channel number of $M_b$. The final adaptive fusion feature $M_{att}$ from the blurry image and events can be obtained by concatenating the $\tilde{M}_{b}$ and $\tilde{M}_{fu}$. The calculation of the adaptive fusion feature $M_{att}$ is applied to every branch with three resolutions. After obtaining adaptive fusion features $M_{att}^{R}$, a Multi-Scale Decoder (MSD) module is applied for feature decoding and enhancement and generates the decoded features $\{M_{md}^{1},M_{esr}^{1}\}=\operatorname{Conv}(M_{att}^{1} \oplus M_{att}^{2 \uparrow} \oplus M_{att}^{4 \uparrow \uparrow})$ for MD and ESR, which $(\cdot)^{\uparrow}$ denotes the combination of upscaling and convolutional layers, as shown in \cref{framework}.  

\begin{figure*}
\centering
\includegraphics[width=.98\textwidth]{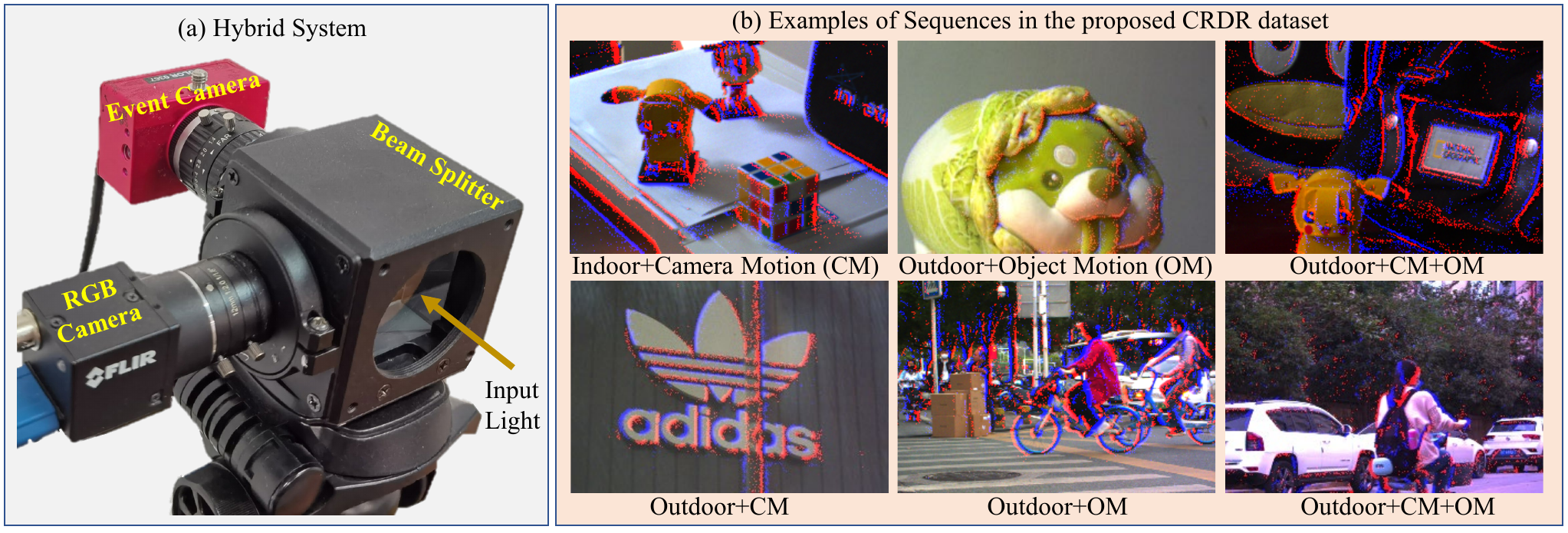}
\caption{(a) The illustration of a beam splitter-based hybrid system for building the CRDR. (b) Sequence samples of the proposed CRDR dataset.}
\label{hybrid_system}
\vspace{-1em}
\end{figure*}

\begin{table}[th]
\small
\centering
\caption{Details about the proposed CRDR dataset. In total, the dataset consists of 76 sequences for both indoor and outdoor scenarios.}
\vspace{-.5em}
\begin{tabular}{lllllccccc}
\hline
{Split} & {Scenes} & {\#Seq} & {\#Events (M)} & {\#Pairs} \\ \hline
{Train} & Buildings & 4 & 23.43 & 299 \\
& Downstairs & 2 & 18.76 & 157 \\
& Campus & 18 & 93.46 & 1153 \\
& Pedestrians & 11 & 31.42 & 528 \\
& Streets & 15 & 59.33 & 978 \\
& Indoor & 5 & 35.16 & 387 \\ \cline{3-5}
& & \textbf{55} & \textbf{261.56} & \textbf{3502} \\ \hline
{Test} & Buildings & 2 & 8.54 & 113 \\
& Downstairs & 1 & 6.52 & 39 \\
& Campus & 7 & 33.71 & 330 \\
& Pedestrians & 3 & 22.77 & 138 \\
& Streets & 5 & 38.51 & 205 \\
& Indoor & 3 & 5.97 & 57 \\ \cline{3-5}
& & \textbf{21} & \textbf{116.02} & \textbf{882} \\  \hline
{Total} & & \textbf{76} & \textbf{377.58} & \textbf{4384} \\
\hline
\end{tabular}
\vspace{-0.em}
\label{tab_crdr}
\end{table}

\subsubsection{Cross-Interaction Prediction}
The Cross-Interaction Prediction (CIP) module is designed to further enhance the intermediate feature map by integrating cross-task representations, and the details of CIP are shown in \cref{framework}. Specifically, the predicted sharp images can be computed by 
\begin{equation}\label{CIP_I}
    \mathbf{I}^{1}=o(\operatorname{Conv}(M_{md}^{1})\oplus \operatorname{Conv}(M_{esr}^{1}))+{\mathbf{B}^{1}},
\end{equation}
where we add the original blurry image to generate the final predicted sharp image $\mathbf{I}^{1}$. For final HR event prediction, the formula can be defined as
\begin{equation}\label{CIP_E}
    \tilde{E}^{1}(t)=o(\operatorname{Conv}(M_{esr}^{1})\oplus \operatorname{Conv}(M_{md}^{1})\oplus E_{\Delta{t}}^{1}(t)),
\end{equation}
which concatenates the time-dependent feature $E_{\Delta{t}}^{1}(t)$ to assist in generating HR events at timestamps $t$.

\subsection{Optimization}
\noindent\textbf{The loss of Motion Deblurring.} We supervise the estimation of latent sharp images with the $\ell_1$ norm between the predicted sharp images $\mathbf{I}(t)$ and the ground truth sharp images $\hat{\mathbf{I}}(t)$, which is defined as
\begin{equation}\label{deblur_loss}
    \mathcal{L}_{md}=\Vert \mathbf{I}(t)-\hat{\mathbf{I}}(t)\Vert{_1}.
\end{equation}
Training CZ-Net at the referenced timestamps with the corresponding ground-truth sharp images is straightforward. When it comes to training at non-referenced timestamps, it can be challenging due to the lack of reference sharp images. As a result, a video interpolation method, \ie, RIFE~\cite{huang2022rife}, is applied to generate reference sharp images at non-referenced timestamps, enabling the supervision of the latent restoration of those non-referenced timestamps. 

\noindent\textbf{The loss of Event Super-Resolving.} Similarly, the event super-resolving task is also supervised by the $\ell_1$ norm between the predicted HR event $\tilde{E}^{1}_{t}$ and the ground truth HR event $\hat{E}^{1}_{t}$, which can be defined as
\begin{equation}\label{event_loss}
    \mathcal{L}_{esr}=\Vert \tilde{E}^{1}(t)-\hat{E}^{1}(t)\Vert{_1}.
\end{equation}
Thus, the overall loss function for CZ-Net is defined as
\begin{equation}\label{all_loss}
    \mathcal{L}_{total}=\mathcal{L}_{md}+\alpha\mathcal{L}_{esr}+\beta\mathcal{L}_{att},
\end{equation}
where $\alpha,\beta$ are the balancing parameters.

\section{CRDR Dataset}\label{crdr-dataset}
To our knowledge, no publicly released dataset is available yet for uEDSR task with Low-Resolution (LR) events. It motivates us to build a new dataset for Cross-Resolution Deblurring and Resolving (CRDR) containing paired HR {\it sharp-blurry} images and the corresponding {\it HR-LR} event streams. 

\noindent\textbf{System Setup.} To simultaneously collect images and events, we build a hybrid camera system composed of an LR DAVIS346 event camera of resolution 346$\times$260 and an HR FLIR BlackFly U332S4 RGB camera of resolution 2048$\times$1536 working at a frame rate of 118 FPS, as shown in \cref{hybrid_system} (a). A beam splitter connects two cameras to achieve minimum spatial parallax between RGB frames and events. Since two cameras provide vision perceptions of different modalities and spatial resolutions, calibrations in both spatial and temporal domains are essential to ensure alignments between collected RGB frames and events.

\noindent\textbf{Spatial and Temporal Calibration.} The spatial and temporal calibration ensures that LR events and HR RGB images are time synchronized and share the same field of view. 
We employ the Active Pixel Sensor (APS) frames output from the DAVIS346 event camera to achieve spatial calibration since APS frames and events share the same light path and spatial resolution. The homography between two cameras can be estimated to achieve the geometric calibration, where the LR APS frames and HR RGB images are matched using SIFT~\cite{lowe2004distinctive} and RANSAC~\cite{fischler1981random}.
Regarding temporal synchronization, we only have to estimate the time offsets between two cameras by searching the timestamps with the highest values of structural similarity~\cite{wang2004image} between stacked LR event frames and the gradient maps of downsampled HR RGB frames. \cref{calibration} illustrates a typical example of the calibration results.
\begin{table*}[th]
\small
\renewcommand{\arraystretch}{1.0}
\centering
\caption{Quantitative comparisons of the proposed CZ-Net to the state-of-the-art motion deblurring methods on GoPro and CRDR datasets. The best performance is in \textcolor{red}{\textbf{bold}} and the second best is \textcolor{blue}{\underline{underlined}}. The symbol / denotes infeasible to reconstruct sharp sequences.
}
\begin{tabular}{cccccccccccccccccc}
\hline
\multirow{3}{*}{Methods} & \multirow{3}{*}{Color} & \multicolumn{5}{c}{Single frame restoration}  &  & \multicolumn{5}{c}{7 frames restoration}
\\ \cline{3-7} \cline{9-13} 
  & & \multicolumn{2}{c}{GoPro} &   & \multicolumn{2}{c}{CRDR} &  & \multicolumn{2}{c}{GoPro} &   & \multicolumn{2}{c}{CRDR} \\ \cline{3-4} \cline{6-7} \cline{9-10} \cline{12-13} 
 & & 
  \multicolumn{1}{c}{PSNR$\uparrow$} &
  \multicolumn{1}{c}{SSIM$\uparrow$} &
   &
  \multicolumn{1}{c}{PSNR$\uparrow$} &
  \multicolumn{1}{c}{SSIM$\uparrow$} &
   &
  PSNR$\uparrow$ &
  SSIM$\uparrow$ &
   &
  PSNR$\uparrow$ &
  SSIM$\uparrow$ \\ \hline

DeblurGanV2~\cite{kupyn2019deblurgan} & \Checkmark &30.98 & 0.9176 & & 23.66 & 0.6973 & & / & / & & /  & /  \\
MPRNet~\cite{Zamir2021MPRNet} & \Checkmark &
32.66 & 0.9590 & & 24.36 & 0.7553 & & / & / & & /  & /  \\
MotionETR~\cite{zhang2021exposure} &\Checkmark & 29.72 & 0.9305  & & 21.46 & 0.6818 & & 23.37 & 0.7328 & & 20.21 & 0.6400 \\
\hline
BL+EF-Net~\cite{sun2022event} &  \Checkmark & 24.38 & 0.7963  & & 22.84 & 0.7083 & & / & / & & / & / \\
EZ~\cite{duan2021eventzoom}+EF-Net~\cite{sun2022event} &  \Checkmark & 26.58 & 0.8440  & & 22.83 & 0.7181 & & / & / & & / & / \\
BL+E-CIR~\cite{song2022cir} &  \XSolidBrush & 21.39 & 0.7907 & & 14.34 & 0.5024 & & 21.28 & 0.7852 & & 14.21 & 0.4988 \\
EZ~\cite{duan2021eventzoom}+E-CIR~\cite{song2022cir} &  \XSolidBrush & 20.31 & 0.7231  & & 13.40 & 0.4535 & & 20.11 & 0.7146 & & 13.31 & 0.4406 \\
eSL-Net$_{4\times}$~\cite{wang2020event} &  \XSolidBrush & 22.29 & 0.6790  & & 14.32 & 0.5318 & & 21.69 & 0.6631 & & 13.41 & 0.4811 \\
RED~\cite{xu2021motion}+DASR~\cite{wang2021unsupervised}  & \XSolidBrush & 22.26   & 0.7210  & & 13.45 & 0.4761 & & 21.24 & 0.6720 & & 13.36 & 0.4741 \\
BL+eSL-Net~\cite{wang2020event}  & \XSolidBrush & 22.19 & 0.7255  & & 13.86 & 0.4784 & & 20.38 & 0.6946 & & 13.22  & 0.4591 \\
BL+RED~\cite{xu2021motion} & \XSolidBrush & 25.79 & 0.8514 & & 18.75 & 0.6494 & & 24.18 & 0.8190 & & 16.49 & 0.6276 \\
EZ~\cite{duan2021eventzoom}+eSL-Net~\cite{wang2020event} & \XSolidBrush & 20.52 & 0.6673  & & 14.35 & 0.5281 & & 21.64 & 0.7109 & & 14.18  & 0.5201 \\
EZ~\cite{duan2021eventzoom}+RED~\cite{xu2021motion} & \XSolidBrush & 25.51 & 0.8209  & & 22.89 & 0.6954 & & 22.60 & 0.7582 & & 19.55 & 0.6203 \\
\hline
CZ-Net-G  & \XSolidBrush & \textcolor{red}{\textbf{35.00}}  & \textcolor{red}{\textbf{0.9659}}  & & \textcolor{blue}{\underline{26.28}} & \textcolor{blue}{\underline{0.7407}} & & \textcolor{blue}{\underline{31.61}} & \textcolor{blue}{\underline{0.9302}} & & \textcolor{blue}{\underline{25.01}} & \textcolor{blue}{\underline{0.7274}} \\
CZ-Net-C  & \Checkmark &  \textcolor{blue}{\underline{34.74}}  & \textcolor{blue}{\underline{0.9642}}  & & \textcolor{red}{\textbf{26.93}} & \textcolor{red}{\textbf{0.7835}} & & \textcolor{red}{\textbf{31.78}} & \textcolor{red}{\textbf{0.9322}} & & \textcolor{red}{\textbf{25.52}} & \textcolor{red}{\textbf{0.7693}} \\
 \hline
\end{tabular}
\vspace{-1em}
\label{tab:qua_exp}
\end{table*}

\def\imwidth{0.498}
\def\cimwid{0.14}

\def\zuoxia{(-0.3,-0.9)}
\def\youshang{(0.2,0.45)}

\def\ssyy{(-0.8,-0.85)}
\def\ssizz{0.5cm}
\def\sswidth{0.245\textwidth}
\def\ssmag{3}
\def\scc{(2.12,1.4)}

\begin{figure}[!htp]
\footnotesize
	\centering

    	\begin{minipage}[t]{\imwidth\linewidth}
    		\centering
			\begin{tikzpicture}[spy using outlines={rectangle,green,magnification=\ssmag,size=\ssizz},inner sep=0]
				\node {\includegraphics[width=\linewidth]{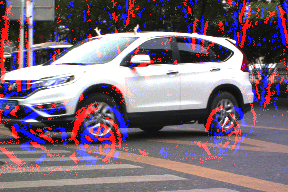}};
				\end{tikzpicture}
    (a) Misaligned Scenes \vspace{0.5em}
    	\end{minipage}%
    \hfill
    	\begin{minipage}[t]{\imwidth\linewidth}
    		\centering
    		\begin{tikzpicture}[spy using outlines={rectangle,green,magnification=\ssmag,size=\ssizz},inner sep=0]
				\node {\includegraphics[width=\linewidth]{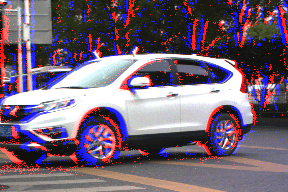}};
				\end{tikzpicture}
			(b) After Global Alignment \vspace{0.5em}
    	\end{minipage}%
    \vspace{-0.5em}
	\caption{Alignment of standard frames with events. Aggregated events (red is positive, and blue is negative) are overlain with the standard frame.}
	\label{calibration}
\end{figure}

\noindent\textbf{Data Collecting.} 
As shown in \cref{tab_crdr}, we collect a total of 76 sequences comprised of both image sequences and event streams with diversity in terms of scenes, \ie, indoor and outdoor, and motion types, \ie, camera and object motion. Some representative examples are visualized in~\cref{hybrid_system} (b).
The sharp HR images are captured with slow and stable camera movement to avoid motion blur. Furthermore, to preserve maximum effectiveness information after spatial and temporal calibration, we crop the RGB frames and events to 768$\times$1152 and 192$\times$288, respectively, by taking the center of them as the basement. 
Motion blurs are synthesized by averaging adjacent 7 HR sharp frames, resulting in 4384 pairs with blurred and sharp images, and the HR-Event streams are synthesized using the event simulator~\cite{rebecq2018esim}. We randomly select 55 sequences as the training set and the rest as the testing set. The data-splitting details of our CRDR dataset are summarized in~\cref{tab_crdr}.

\section{Experiments and Analysis}
In this section, the performance of our CZ-Net is evaluated on both MD and ESR tasks. We first present the experimental settings in~\cref{ex1}, including details on datasets and implementations. Afterward, quantitative and qualitative comparisons of our CZ-Net are made to the state-of-the-art MD methods on restorations of a single latent sharp image in~\cref{ex2} and a continuous-time video sequence in~\cref{ex3}. We further evaluate the performance of CZ-Net on the ESR task in~\cref{ex4}. Ablations on the effectiveness of network architecture and loss functions are finally presented in \cref{ex5}.

\input{figs/com_exp_pred1.tex}

\subsection{Experimental Settings}
\label{ex1}
\noindent\textbf{Datasets.} Two datasets are employed for network training and evaluation, \ie, the synthetic dataset simulated over GoPro~\cite{nah2017deep} and the real-world CRDR dataset built in this paper. Both datasets contain paired {\it HR-LR} events and HR {\it sharp-blurry} images. We have presented the details of our CRDR dataset in \cref{crdr-dataset}. Regarding the synthetic dataset,  we build it upon the GoPro dataset~\cite{nah2017deep}, which is widely adopted to evaluate the MD performance~\cite{zhang2021exposure,Zamir2021MPRNet} and recently used to benchmark the Event-based MD (EMD) task~\cite{xu2021motion,zhang2022unifying,pan2019bringing}. Specifically, we first downsample the HR sharp images from the GoPro dataset and then simulate the LR events by an open-sourced event simulator, \ie, ESIM~\cite{rebecq2018esim}, from the downsampled LR images. At the same time, we synthesize the HR blurry images by averaging the adjacent HR sharp images. The splitting of the training and testing for the synthetic dataset is identical to the official open-sourced GoPro dataset.

\noindent\textbf{Training Details and Evaluation Metrics.}
The proposed CZ-Net is implemented by PyTorch and trained on an NVIDIA GeForce RTX 3090 for 200 epochs with 4 batch sizes by default. Furthermore, we augment the training data by flipping (horizontal, vertical, and horizontal-vertical) and rotating (random angles ranging from $-10${\textdegree} to $10${\textdegree)} to enhance the robustness of our CZ-Net. We use the ADAM optimizer~\cite{kingma2014adam} with an initial learning rate of $10^{-4}$, and the exponential term decays by $0.98$ for every $5$ epoch. The weighting factors $\alpha$ and $\beta$ in \cref{all_loss} are set to $1$ and the $\Delta{t}$ in \cref{ER} is set to 0.2. We use Structural SIMilarity (SSIM)~\cite{wang2004image} and Peak Signal to Noise Ratio (PSNR) as performance metrics for the MD and ESR tasks following~\cite{xu2021motion,weng2022boosting,duan2021eventzoom}.

\input{figs/com_exp_seq.tex}

\subsection{Performance of Motion Deburring}
We first evaluate our CZ-Net for the MD task.
Qualitative and quantitative comparisons are made to state-of-the-art (SOTA) frame-based MD methods, including DeblurGanV2~\cite{kupyn2019deblurgan}, METR~\cite{zhang2021exposure}, and MPRNet~\cite{Zamir2021MPRNet}, and Event-based MD (EMD) methods, including RED~\cite{xu2021motion}, eSL-Net~\cite{wang2020event}, E-CIR~\cite{song2022cir}, and EF-Net~\cite{sun2022event}. We use their official code and test with the default parameter settings. 
Different from our CZ-Net, existing EMD approaches are implemented by superimposing an individual super-resolution module to bridge the scale gap between LR events and HR images, either in the event domain by BiLinear upsampling (BL) and EventZoom (EZ)~\cite{duan2021eventzoom} or in the image domain by DASR~\cite{wang2021unsupervised}. Specifically, we denote BL/EZ+eSL-Net/RED/EF-Net/E-CIR as ESR-then-EMD approaches and RED+DASR as EMD-then-SR approaches. Note that eSL-Net~\cite{wang2020event} itself can super-resolve the LR blurry image. Thus, we implement it by first downsampling the HR blurry image and then super-resolving the downsampled LR blurry image with the LR events, denoted as eSL-Net$_{4\times}$.

\subsubsection{Results of Single Frame Restoration} \label{ex2} 

\input{figs/com_exp_esr_pred1}
\noindent\textbf{Quantitative results.}
The quantitative results of the single-frame prediction on the GoPro and the proposed CRDR dataset are shown in \cref{tab:qua_exp}. Our CZ-Net outperforms state-of-the-art methods by a large margin. Specifically, CZ-Net achieves 8.3/6.9 dB and 0.158/0.151 improvements in PSNR and SSIM on the GoPro and CRDR datasets. The superior performance of CZ-Net is attributed to its effective utilization of flawed LR events, which still contain essential intra-frame information on motion and texture. Among the EMD methods, our proposed CZ-Net exhibits significant improvements over the two-stage cascaded methods, \ie, BL+EF-Net, BL+E-CIR, RED+DASR, BL+eSL-Net, BL+RED, EZ+EF-Net, EZ+E-CIR, EZ+eSL-Net, and EZ+RED, which validate the effectiveness of our unified framework that can eliminate the accumulated errors between the separated ESR and MD tasks. 

\noindent\textbf{Qualitative results.} We further qualitatively evaluate the performance in the GoPro and CRDR datasets, and qualitative results for visual comparison are shown in \cref{exp_compare_pred1}. On the one hand, the results in the GoPro dataset are shown in the top two rows of \cref{exp_compare_pred1}. The results estimated by frame-based SOTA methods, \ie, MPRNet~\cite{Zamir2021MPRNet} and METR~\cite{zhang2021exposure}, suffer artifacts and distortions, degenerating the overall quality of the deblurred image. Moreover, the two-stage architecture combining the ESR methods accumulates defects in each independent task, leading to severe degeneration of the deblurring results. On the contrary, Our CZ-Net can obtain accurate reconstructions that exhibit the most similar appearances to the ground-truth sharp images. On the other hand, the bottom two rows of \cref{exp_compare_pred1} present qualitative results on the CRDR dataset with {\it real-world} LR events in more challenging scenarios. The MPRNet~\cite{Zamir2021MPRNet} and METR~\cite{zhang2021exposure} fail to restore sharp latent images without the aid of events due to the increased complexities of the scene in motions and textures. Frame-based methods are ineffective in recovering targets detailly, \eg, {\it handrails}. Cascaded event-based methods, \ie, BL+RED, cannot restore sharper contours of targets and suffer suboptimal solutions owing to the error propagation. Our CZ-Net predicts results with sharper edges and smoother surfaces than its competitors, demonstrating the effectiveness of our unifying architecture and the AAE mechanism in suppressing invalid features caused by LR events.

\subsubsection{Results of Video Sequence Restoration} \label{ex3}
Video sequence recovery from a single blurry image is more challenging than single-frame reconstruction. To our knowledge, most of the existing frames-based MD approaches can only predict a single frame, \ie, DeblurGanV2~\cite{Kupyn2018CVPR} and MPRNet~\cite{Zamir2021MPRNet}. Thus, we only compare our proposed CZ-Net with the methods, \ie, METR, EZ+RED, EZ+eSL-Net, BL+eSL-Net, and BL+RED, that can predict latent sharp sequence from a single blurry image. Quantitative and qualitative results are given in \cref{tab:qua_exp} and \cref{exp_compare_seq}, respectively.

Regarding the quantitative results, we evaluate the performance of MD methods concerning seven reconstructed latent sharp images from a single blurry input. On the GoPro dataset, our CZ-Net outperforms other methods regarding SSIM (up to 0.1722 improvement) and PSNR (up to 6.04 dB improvement). On the CRDR dataset, our CZ-Net performs best, validating its effectiveness in handling real-world events.

To explore the superiority of our CZ-Net in recovering sharp latent sequences, we compare BL+RED and METR in reconstructing 15 continuous frames of the CRDR dataset. Since RED and METR can only output 7 latent frames, we cascade these methods and the video frame interpolation method, \ie, RIFT~\cite{huang2022rife} to achieve continuous-time restoration, where 7 deblurred images are interpolated to 15 images. The qualitative results are shown in \cref{exp_compare_seq}. METR can restore the sharp frame at the intermediate exposure time. Still, the effects in the whole temporal domain degrade severely due to the inconsistency between the motion assumption and the scenes with complex motions. Moreover, BL+RED performs with a slight improvement in both spatial and temporal domains compared to METR, thanks to the introduction of LR events. However, its details still suffer artifacts, \eg, halo effects, since deblurring errors might be propagated to the interpolation stage and accumulated to the restorations of the non-referenced timestamps. Our CZ-Net predicts results with sharper edges and smoother inter-frame transitions than the SOTA methods at non-reference and reference timestamps.

\subsection{Performance of Event Super-Resolving} \label{ex4}
The uEDSR is a joint task in which both MD and ESR should be addressed. Therefore, we compare our CZ-Net to the state-of-the-art frame-based SR methods, \ie, SRFBN~\cite{li2019feedback}, and the event-based method, \ie, RecEvSR~\cite{weng2022boosting}. Furthermore, we retrain RecEvSR~\cite{weng2022boosting} on the GoPro and CRDR datasets using the official code with the default parameter setting for a fair comparison.

\subsubsection{Qualitative Results} 
We evaluate the performance qualitatively on the GoPro and CRDR datasets. The results are visualized in \cref{esr_gopro} for the GoPro dataset and \cref{exp_compare_esr_seq} for the CRDR dataset.

\cref{esr_gopro} illustrates the qualitative results on the GoPro dataset under the middle time of the exposure period of the corresponding blurry image. The results of the SRFBN and RecEvSR methods still suffer texture discontinuities and jagged distortions, thus failing to recover the {\it license plate} of the second example in ~\cref{esr_gopro}. On the contrary, our CZ-Net gives precise HR event reconstructions in the spatial domain, exhibiting the most similar appearances to ground-truth HR events.

\input{figs/com_exp_esr_seq}

\begin{table}[!t]
\centering
\small
\renewcommand{\arraystretch}{1.}
\caption{Quantitative comparisons of ESR in terms of E2VID results on the GoPro dataset, where SSIM ($\uparrow$) is adopted as the metric. \textcolor{red}{\textbf{Bold}} and \textcolor{blue}{\underline{underlined}} numbers represent the best and second-best performance.}
\begin{tabular}{c|cccc}
\hline
Sequence & $\operatorname{BL}$ &  $\operatorname{SRFBN}$  & $\operatorname{RecEvSR}$ & CZ-Net
\\ 
\hline
\#0  & 0.2822  &  \textcolor{blue}{\underline{0.3014}} & 0.2751  & \textcolor{red}{\textbf{0.3638}} 
\\
\#1  &  0.1910 & \textcolor{blue}{\underline{0.1945}}  &  0.1693  & \textcolor{red}{\textbf{0.1963}}
\\
\#2  &  0.3001  & \textcolor{blue}{\underline{0.3286}}  &  0.2766 &  \textcolor{red}{\textbf{0.3676}}
\\
\#3 & 0.2901 &  \textcolor{blue}{\underline{0.3179}}  &  0.2473  &  \textcolor{red}{\textbf{0.3815}}
\\
\#4 &  0.2699 &  \textcolor{blue}{\underline{0.2975}} & 0.2688  & \textcolor{red}{\textbf{0.3421}} 
\\
\#5 &  0.4277  &  \textcolor{blue}{\underline{0.4422}}  &  0.3398 &  \textcolor{red}{\textbf{0.4574}}
\\
\#6 & 0.3783  &  \textcolor{blue}{\underline{0.4033}}  &  0.2911 & \textcolor{red}{\textbf{0.4340}}
\\
\#7  &  0.4239  &  0.4423 & \textcolor{blue}{\underline{0.4749}} & \textcolor{red}{\textbf{0.5202}}
\\
\#8  &  0.4683  &  \textcolor{blue}{\underline{0.4920}}  &  0.3746 & \textcolor{red}{\textbf{0.5009}}
\\
\#9  &  \textcolor{blue}{\underline{0.4409}}  &  \textcolor{red}{\textbf{0.4595}}  & 0.3119  & 0.4380
\\
\#10  &  0.2237  &  \textcolor{blue}{\underline{0.2419}}  &  0.2190 & \textcolor{red}{\textbf{0.3155}}
\\ \hline
Average   &  0.3394  &   \textcolor{blue}{\underline{0.3598}}  &  0.2955 & \textcolor{red}{\textbf{0.3930}}
\\ \hline
\end{tabular}
\label{tab:esr_e2vid}
\vspace{-.5em}
\end{table}
\def\imwidth{0.33}
\def\cimwid{0.14}

\def\zuoxia{(-0.3,-0.9)}
\def\youshang{(0.2,0.45)}

\def\ssyy{(-0.8,-0.85)}
\def\ssizz{0.5cm}
\def\sswidth{0.245\textwidth}
\def\ssmag{3}
\def\scc{(2.12,1.4)}

\begin{figure}[!htb]
\footnotesize
	\centering
\begin{minipage}[t]{\imwidth\linewidth}
    		\centering
			\begin{tikzpicture}[spy using outlines={rectangle,green,magnification=\ssmag,size=\ssizz},inner sep=0]
				\node {\includegraphics[width=\linewidth]{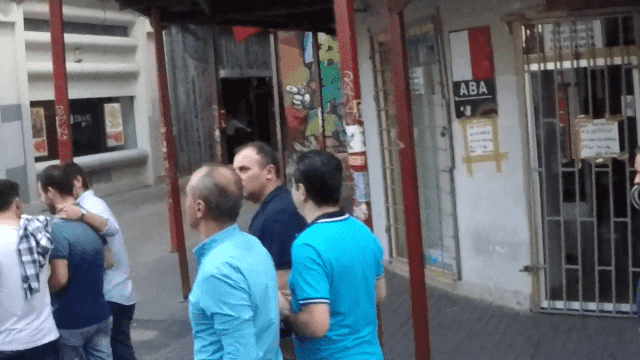}};
				\end{tikzpicture}
            (a) Frame 
            \\
                        \vspace{.5mm}
            \begin{tikzpicture}[spy using outlines={green,magnification=\ssmag,size=\ssizz},inner sep=0]
				\node {\includegraphics[width=\linewidth,trim={0 0 0 0},clip,cframe=black .0001mm]{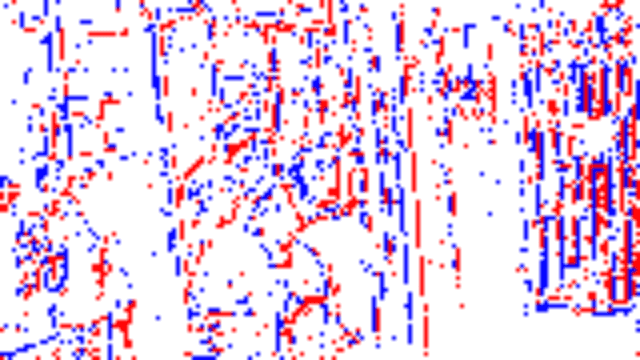}};
				\end{tikzpicture}
    (d) LR Events
			 \vspace{0.5em}
\end{minipage}%
    \hfill
\begin{minipage}[t]{\imwidth\linewidth}
    		\centering
    	    \begin{tikzpicture}[spy using outlines={green,magnification=\ssmag,size=\ssizz},inner sep=0]
				\node {\includegraphics[width=\linewidth]{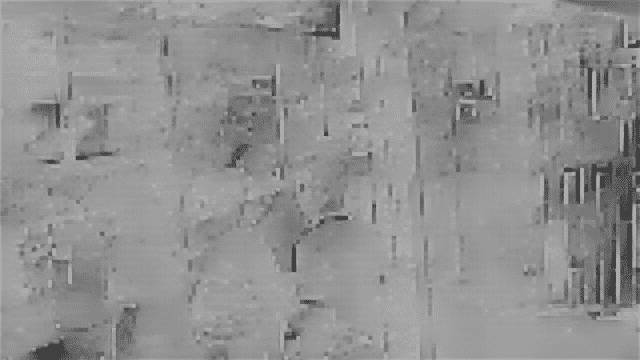}};
				\end{tikzpicture}
    (b) BL
   \\
      \vspace{.5mm}
   \begin{tikzpicture}[spy using outlines={green,magnification=\ssmag,size=\ssizz},inner sep=0]
				\node {\includegraphics[width=\linewidth]{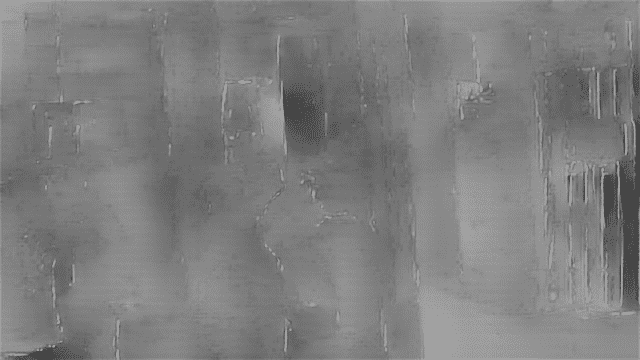}};
				\end{tikzpicture}
    (e) SRFBN
			\vspace{0.5em}
\end{minipage}%
    \hfill
\begin{minipage}[t]{\imwidth\linewidth}
    		\centering
    	    \begin{tikzpicture}[spy using outlines={green,magnification=\ssmag,size=\ssizz},inner sep=0]
				\node {\includegraphics[width=\linewidth]{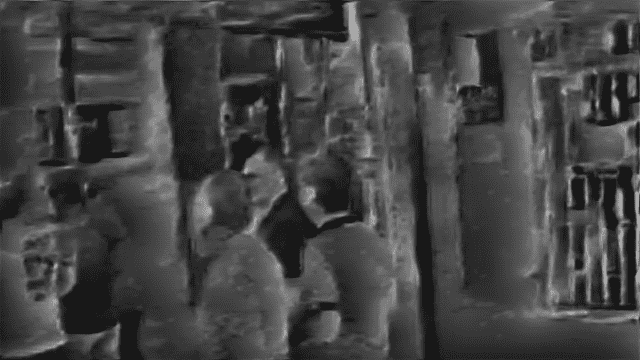}};
				\end{tikzpicture}
    (c) RecEvSR
   \\
      \vspace{.5mm}
            \begin{tikzpicture}[spy using outlines={green,magnification=\ssmag,size=\ssizz},inner sep=0]
				\node {\includegraphics[width=\linewidth]{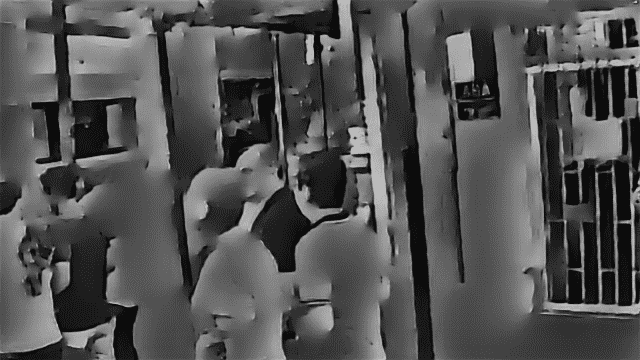}};
				\end{tikzpicture}
    (f) CZ-Net
			\vspace{0.5em}
\end{minipage}%
    \hfill
  \caption{Visual comparisons of video reconstruction among BiLinear (BL), SRFBN~\cite{li2019feedback}, RecEVSR~\cite{weng2022boosting} and ours CZ-Net on the GoPro dataset.}
  \vspace{-.5em}
\label{esr_e2vid_tab}
  \end{figure}

\cref{exp_compare_esr_seq} visualizes the reconstruction of 11 sequences at timestamps with equal intervals to demonstrate the superiority of our CZ-Net. In addition to the advantages in terms of texture details in the spatial domain, our CZ-Net also has significant advantages in terms of temporal accuracy and noise robustness compared to RecEvSR as illustrated in the last three rows of \cref{exp_compare_esr_seq}.

\subsubsection{Quantitative Results on Downstream Applications} 
We investigate the performance of BiLinear (BL) upsampling, SRFBN~\cite{li2019feedback}, RecEVSR~\cite{weng2022boosting}, and our CZ-Net on video reconstruction application over the GoPro dataset. The E2VID~\cite{rebecq2019high} is chosen as the benchmark algorithm for event-to-video reconstruction with the evaluation metric of SSIM. The numerical results are presented in \cref{tab:esr_e2vid}. Our CZ-Net outperforms others in most sequences, demonstrating its superior performance, and our results consistently exhibit the best average performance across all sequences. Moreover, the visual comparison results are displayed in \cref{esr_e2vid_tab}. These results demonstrate that our CZ-Net captures fine perceptual details while the comparison methods (BL, SRFBN, and RecEvSR) generate noticeable artifacts.

\input{figs/com_exp_ablation_module}
\begin{table}[!t]
\centering
\small
\renewcommand{\arraystretch}{1.0}
\caption{Ablation study of MBEF, AAE, and CIP in our method on the CRDR dataset. \textcolor{red}{\textbf{Bold}} and \textcolor{blue}{\underline{underlined}} numbers represent the best and the second-best performance.}
\begin{tabular}{c|ccc|cc}
\hline
Case & $\operatorname{MBEF}$ &  $\operatorname{AAE}$  & $\operatorname{CIP}$ & PSNR$\uparrow$ & SSIM$\uparrow$ 
\\ 
\hline
\#0  &  $\checkmark$  &   &   & 24.40  &  0.7411 
\\
\#1    &   &    $\checkmark$    &    &  25.39  & 0.7593
\\
\#2    &    &     & $\checkmark$   &  25.59 &  0.7622
\\
\#3   &   $\checkmark$   &   $\checkmark$   &    & 25.67  & 0.7640
\\
\#4    &     $\checkmark$   &     & $\checkmark$ & \textcolor{blue}{\underline{26.34}} & \textcolor{blue}{\underline{0.7750}}
\\
\#5     &     & $\checkmark$      &    $\checkmark$    &  
25.74 & 0.7654
\\
\#6    & $\checkmark$      & $\checkmark$    &  $\checkmark$      & 
\textcolor{red}{\textbf{26.93}} & \textcolor{red}{\textbf{0.7836}}    
\\ \hline
\end{tabular}
\label{tab:ablation}
\vspace{-1mm}
\end{table}
\begin{table}[!t]
\centering
\small
\renewcommand{\arraystretch}{1.0}
\caption{Ablation study supervision on the CRDR dataset. \textcolor{red}{\textbf{Bold}} numbers represent the best performance.}
\begin{tabular}{c|ccc|cc}
\hline
Ex. & $\mathcal{L}_{md}$ &  $\mathcal{L}_{esr}$  & $\mathcal{L}_{att}$ & PSNR$\uparrow$ & SSIM$\uparrow$ 
\\ 
\hline
A  &  $\checkmark$  &   &   & 25.43  &  0.7624 
\\
B   &  $\checkmark$ &   $\checkmark$    &    &  25.86  & 0.7653
\\
C   &  $\checkmark$  &     & $\checkmark$   &  26.05 &  0.7712
\\
D   &   $\checkmark$   &   $\checkmark$   &  $\checkmark$  & \textcolor{red}{\textbf{26.93}}  & \textcolor{red}{\textbf{0.7836}}
\\ \hline
\end{tabular}
\label{tab:ablation_loss}
\vspace{-1mm}
\end{table}

\subsection{Ablation Study}
\label{ex5}

In this subsection, we conduct a comprehensive set of ablation studies to investigate the design choices of our method. Firstly, we demonstrate the individual performance contribution of each module in the network architecture (\cref{tab:ablation} and \cref{ablation_modules,fig:feature_map_aae}). Then, we analyze the role of each supervision during the training phase (\cref{tab:ablation_loss} and \cref{fig:ablation_loss}).

\subsubsection{Network Architecture} The proposed network is composed of a Multi-scale Blur-Event Fusion (MBEF) module, an Attention-based Adaptive Enhancement (AAE) module, and a Cross-Interaction Prediction (CIP) module. Ablation studies are conducted on the CRDR dataset with real-world LR events, where seven different experiments are implemented to analyze the effectiveness of each module. Quantitative and qualitative results are shown in \cref{tab:ablation} and \cref{ablation_modules}, respectively.

\input{figs/com_exp_ablation_loss_v3}

First, we remove the MBEF module (Case 1, 2, and 5 in \cref{tab:ablation}) and replace it with a single-scale fusion module by upsampling the LR events followed by convolutional layers. Compared to it, the full CZ-Net utilizes MBEF for scale-variant enhancement by adapting parallel fusion branches at multi-scale levels. Such a fusion mechanism significantly improves PSNR ($1.19$ dB) and SSIM ($0.0182$). Qualitative results in~\cref{ablation_modules} comparing (c) with (g), (d) with (h), and (i) with (e) demonstrate that CZ-Nets with the MBEF module give sharper results than networks without it, validating the effectiveness of the MBEF module.

\begin{figure*}
\centering
\includegraphics[width=.98\textwidth]{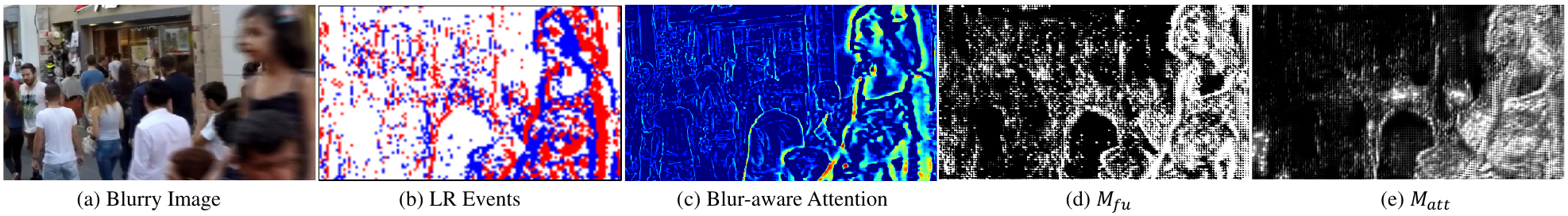}
\caption{Visualization of input (d) and output (e) feature maps of AAE on the GoPro dataset.}
\label{fig:feature_map_aae}
\vspace{-.5em}
\end{figure*}

Moreover, the AAE module is designed to alleviate the distortions caused by the low spatial resolution of the LR events. The deletion of this module (Case 0, 3, and 4) brings performance degeneration, \ie, $0.59$ dB and $0.0086$ in terms of PSNR and SSIM, and the corresponding visualization results shown in \cref{ablation_modules} (c), (f), and (g) suffer severe artifacts. Additionally, the feature visualizations of the AAE module are depicted in \cref{fig:feature_map_aae}. Owing to the tremendous coupled distortions and noise in LR events (b), the area in the event feature map (d) and sharp pixels (a) is invalid and should be suppressed. Compared with the input feature map (d), the AAE module's output feature map (e) effectively reduces the confidence in such valid areas, validating its ability for invalid feature suppression.

\input{figs/com_exp_ablation_blur_loss}

Furthermore, we exclude the CIP module (cases 0, 1, and 3 in \cref{tab:ablation}) and substitute it with a pair of conventional dual-branch convolutional layers with an equivalent number of parameters. In contrast, the full CZ-Net integrates CIP for frame-event complementary enhancement by adapting a feature cross-fertilization strategy. This integration substantially improves PSNR ($1.36$ dB) and SSIM ($0.0196$), quantitatively validating our approach. Qualitatively, illustrated in \cref{ablation_modules} and compared between instances (f) and (c), (h) and (e), as well as (i) and (g), CZ-Net with the CIP module consistently yield sharper results than their module-less counterparts. This visually attests to the effectiveness of the CIP module within the CZ-Net framework.    


\subsubsection{Necessity of Loss Combination} From the absolute difference to the ground-truth sharp image and the metric performance (\ie, PSNR$\uparrow$/SSIM$\uparrow$) shown in \cref{fig:ablation_loss}, we can observe that removing either $\mathcal{L}_{esr}$ or $\mathcal{L}_{att}$ leads to a degradation of the prediction of the results, indicating their contribution to restoring the sharp latent image. Specifically, $\mathcal{L}_{esr}$ introduces the intra-frame information, \ie, sharp texture, embedded in the HR event streams, which can guide the clearer details recovering for the MD task. $\mathcal{L}_{att}$ provides blur-aware attention supervision to guide AAEs to alleviate the distortions embedded in the LR events and thus is also helpful in the entire MD task. Moreover, experiments are conducted to combine the above two loss functions to validate the effectiveness further. In general, \cref{tab:ablation} and \cref{fig:ablation_loss} show that combining all these losses leads to the smallest absolute error, validating the necessity of recovering latent sharp images with $\mathcal{L}_{esr}$ and $\mathcal{L}_{att}$ simultaneously. Furthermore, through qualitative ablations of $\mathcal{L}_{md}$ presented in \cref{exp_ablation_wo_loss_blur}, it becomes apparent that excluding the MD loss, $\mathcal{L}_{md}$, results in a deterioration in the HR events prediction, exemplified by the instance of the {\it wheel} in \cref{exp_ablation_wo_loss_blur}, indicating their contribution to furnish sharp details that aid in the recovery of HR event details.

\subsection{Complexity to Performance Analysis}
\begin{figure}
\centering
\vspace{-.8em}
\includegraphics[width=.45\textwidth]{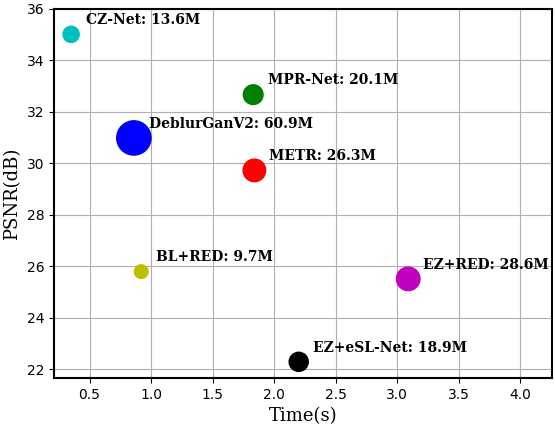}
\caption{Performance to complexity diagram on the GoPro dataset. The inference time is evaluated over the GoPro dataset by feeding blurry images of 640$\times$360, where the model in the top left corner has the highest efficiency and effectiveness. The size of the blobs is proportional to the number of network parameters.}
\label{fig:infertime}
\vspace{-.2em}
\end{figure}

We further evaluate the complexity of our CZ-Net and SOTA MD and ESR methods on the synthetic GoPro dataset by feeding a blurry image with $640\times360$ and LR event stream with $160\times90$ and implementing them on an NVIDIA GeForce RTX 3090 GPU. The comparison results on the complexity-to-performance diagram are given in \cref{fig:infertime}. Our CZ-Net achieves dominant performance in terms of PSNR ($35.00$ dB) but at the expense of the smallest time consumption ($\approx 0.35$ s), validating its superiority in efficiency and effectiveness. Furthermore, the model size regarding the number of parameters is also indicated in \cref{fig:infertime}, where our CZ-Net is the second smallest network and has only $13.6$ M parameters, validating its memory efficiency.

\section{Conclusion}
In this paper, we propose a novel CZ-Net to simultaneously achieve the sharp latent sequence restoration and the HR event stream reconstruction from a single blurry image and its concurrent LR event stream. Specifically, a multi-scale blur event fusion architecture is presented to achieve cross-enhancement by adequately utilizing scale-variant and fusing cross-modality data. Attention-based adaptive enhancement and cross-interaction prediction modules are designed to alleviate the coupled distortions embedded in LR events and reinforce the final results through the prior information. Furthermore, we proposed a hybrid system to build a new dataset containing high-resolution, sharp sequences and the corresponding {\it real-world} LR event streams to narrow the real-to-synthetic gap. Extensive qualitative and quantitative experiments on synthetic and real-world datasets demonstrate that the proposed CZ-Net achieves state-of-the-art deblurring and event super-resolving performance.

\bibliographystyle{ieee_fullname}
\bibliography{egbib}

\begin{thebibliography}{10}\itemsep=-1pt

\bibitem{arjomand2017deep}
Siavash Arjomand~Bigdeli, Matthias Zwicker, Paolo Favaro, and Meiguang Jin.
\newblock Deep mean-shift priors for image restoration.
\newblock {\em NeurIPS}, 30, 2017.

\bibitem{chan1998total}
Tony~F. Chan and Chiu-Kwong Wong.
\newblock Total variation blind deconvolution.
\newblock {\em IEEE TIP}, 7(3):370--375, 1998.

\bibitem{chen2019blind}
Liang Chen, Faming Fang, Tingting Wang, and Guixu Zhang.
\newblock Blind image deblurring with local maximum gradient prior.
\newblock In {\em CVPR}, pages 1742--1750, 2019.

\bibitem{cho2011handling}
Sunghyun Cho, Jue Wang, and Seungyong Lee.
\newblock Handling outliers in non-blind image deconvolution.
\newblock In {\em ICCV}, pages 495--502, 2011.

\bibitem{cho2021rethinking}
Sung-Jin Cho, Seo-Won Ji, Jun-Pyo Hong, Seung-Won Jung, and Sung-Jea Ko.
\newblock Rethinking coarse-to-fine approach in single image deblurring.
\newblock In {\em CVPR}, pages 4641--4650, 2021.

\bibitem{dong2017blind}
Jiangxin Dong, Jinshan Pan, Zhixun Su, and Ming-Hsuan Yang.
\newblock Blind image deblurring with outlier handling.
\newblock In {\em ICCV}, pages 2478--2486, 2017.

\bibitem{duan2021eventzoom}
Peiqi Duan, Zihao~W Wang, Xinyu Zhou, Yi Ma, and Boxin Shi.
\newblock Eventzoom: Learning to denoise and super resolve neuromorphic events.
\newblock In {\em CVPR}, pages 12824--12833, 2021.

\bibitem{fergus2006removing}
Rob Fergus, Barun Singh, Aaron Hertzmann, Sam~T. Roweis, and William~T. Freeman.
\newblock Removing camera shake from a single photograph.
\newblock In {\em ACM SIGGRAPH}, pages 787--794, 2006.

\bibitem{fischler1981random}
Martin~A Fischler and Robert~C Bolles.
\newblock Random sample consensus: a paradigm for model fitting with applications to image analysis and automated cartography.
\newblock {\em Communications of the ACM}, 24(6):381--395, 1981.

\bibitem{Guillermo2020PAMI}
Guillermo Gallego, Tobi Delbruck, Garrick Orchard, Chiara Bartolozzi, Brian Taba, Andrea Censi, Stefan Leutenegger, Andrew~J. Davison, Jorg Conradt, Kostas Daniilidis, and Davide Scaramuzza.
\newblock Event-based vision: A survey.
\newblock {\em IEEE TPAMI}, 44(1):154--180, 2020.

\bibitem{gao2019dynamic}
Hongyun Gao, Xin Tao, Xiaoyong Shen, and Jiaya Jia.
\newblock Dynamic scene deblurring with parameter selective sharing and nested skip connections.
\newblock In {\em CVPR}, pages 3848--3856, 2019.

\bibitem{han2021evintsr}
Jin Han, Yixin Yang, Chu Zhou, Chao Xu, and Boxin Shi.
\newblock Evintsr-net: Event guided multiple latent frames reconstruction and super-resolution.
\newblock In {\em ICCV}, pages 4882--4891, 2021.

\bibitem{haoyu2020learning}
Chen Haoyu, Teng Minggui, Shi Boxin, Wang Yizhou, and Huang Tiejun.
\newblock Learning to deblur and generate high frame rate video with an event camera.
\newblock {\em arXiv preprint arXiv:2003.00847}, 2020.

\bibitem{huang2022rife}
Zhewei Huang, Tianyuan Zhang, Wen Heng, Boxin Shi, and Shuchang Zhou.
\newblock Real-time intermediate flow estimation for video frame interpolation.
\newblock In {\em ECCV}, 2022.

\bibitem{jiang2019mixed}
Zhuangyi Jiang, Pengfei Xia, Kai Huang, Walter Stechele, Guang Chen, Zhenshan Bing, and Alois Knoll.
\newblock Mixed frame-/event-driven fast pedestrian detection.
\newblock In {\em ICRA}, pages 8332--8338, 2019.

\bibitem{jing2021turning}
Yongcheng Jing, Yiding Yang, Xinchao Wang, Mingli Song, and Dacheng Tao.
\newblock Turning frequency to resolution: Video super-resolution via event cameras.
\newblock In {\em CVPR}, pages 7772--7781, 2021.

\bibitem{kingma2014adam}
Diederik~P Kingma and Jimmy Ba.
\newblock Adam: A method for stochastic optimization.
\newblock In {\em ICLR}, 2015.

\bibitem{Kupyn2018CVPR}
Orest Kupyn, Volodymyr Budzan, Mykola Mykhailych, Dmytro Mishkin, and Jiri Matas.
\newblock Deblurgan: Blind motion deblurring using conditional adversarial networks.
\newblock In {\em CVPR}, pages 8183--8192, 2018.

\bibitem{kupyn2019deblurgan}
Orest Kupyn, Tetiana Martyniuk, Junru Wu, and Zhangyang Wang.
\newblock Deblurgan-v2: Deblurring (orders-of-magnitude) faster and better.
\newblock In {\em ICCV}, pages 8878--8887, 2019.

\bibitem{li2019super}
Hongmin Li, Guoqi Li, and Luping Shi.
\newblock Super-resolution of spatiotemporal event-stream image.
\newblock {\em Neurocomputing}, 335:206--214, 2019.

\bibitem{li2019feedback}
Zhen Li, Jinglei Yang, Zheng Liu, Xiaomin Yang, Gwanggil Jeon, and Wei Wu.
\newblock Feedback network for image super-resolution.
\newblock In {\em CVPR}, pages 3867--3876, 2019.

\bibitem{lowe2004distinctive}
David~G Lowe.
\newblock Distinctive image features from scale-invariant keypoints.
\newblock {\em IJCV}, 60(2):91--110, 2004.

\bibitem{michaeli2014blind}
Tomer Michaeli and Michal Irani.
\newblock Blind deblurring using internal patch recurrence.
\newblock In {\em ECCV}, pages 783--798, 2014.

\bibitem{nah2017deep}
Seungjun Nah, Tae Hyun~Kim, and Kyoung Mu~Lee.
\newblock Deep multi-scale convolutional neural network for dynamic scene deblurring.
\newblock In {\em CVPR}, pages 3883--3891, 2017.

\bibitem{pan2020high}
Liyuan Pan, Richard Hartley, Cedric Scheerlinck, Miaomiao Liu, Xin Yu, and Yuchao Dai.
\newblock High frame rate video reconstruction based on an event camera.
\newblock {\em IEEE TPAMI}, 44(5):2519--2533, 2022.

\bibitem{pan2019bringing}
Liyuan Pan, Cedric Scheerlinck, Xin Yu, Richard Hartley, Miaomiao Liu, and Yuchao Dai.
\newblock Bringing a blurry frame alive at high frame-rate with an event camera.
\newblock In {\em CVPR}, pages 6820--6829, 2019.

\bibitem{qian2018attentive}
Rui Qian, Robby~T. Tan, Wenhan Yang, Jiajun Su, and Jiaying Liu.
\newblock Attentive generative adversarial network for raindrop removal from a single image.
\newblock In {\em CVPR}, pages 2482--2491, 2018.

\bibitem{rebecq2018esim}
Henri Rebecq, Daniel Gehrig, and Davide Scaramuzza.
\newblock Esim: an open event camera simulator.
\newblock In {\em Conference on Robot Learning}, pages 969--982. PMLR, 2018.

\bibitem{rebecq2019high}
Henri Rebecq, Ren{\'e} Ranftl, Vladlen Koltun, and Davide Scaramuzza.
\newblock High speed and high dynamic range video with an event camera.
\newblock {\em IEEE TPAMI}, 43(6):1964--1980, 2019.

\bibitem{schmidt2013discriminative}
Uwe Schmidt, Carsten Rother, Sebastian Nowozin, Jeremy Jancsary, and Stefan Roth.
\newblock Discriminative non-blind deblurring.
\newblock In {\em CVPR}, pages 604--611, 2013.

\bibitem{song2022cir}
Chen Song, Qixing Huang, and Chandrajit Bajaj.
\newblock E-cir: Event-enhanced continuous intensity recovery.
\newblock In {\em CVPR}, pages 7803--7812, 2022.

\bibitem{sun2022event}
Lei Sun, Christos Sakaridis, Jingyun Liang, Qi Jiang, Kailun Yang, Peng Sun, Yaozu Ye, Kaiwei Wang, and Luc~Van Gool.
\newblock Event-based fusion for motion deblurring with cross-modal attention.
\newblock In {\em ECCV}, pages 412--428, 2022.

\bibitem{szeliski2022computer}
Richard Szeliski.
\newblock {\em Computer vision: algorithms and applications}.
\newblock Springer Nature, 2022.

\bibitem{wang2020event}
Bishan Wang, Jingwei He, Lei Yu, Gui-Song Xia, and Wen Yang.
\newblock Event enhanced high-quality image recovery.
\newblock In {\em ECCV}, pages 155--171, 2020.

\bibitem{wang2021unsupervised}
Longguang Wang, Yingqian Wang, Xiaoyu Dong, Qingyu Xu, Jungang Yang, Wei An, and Yulan Guo.
\newblock Unsupervised degradation representation learning for blind super-resolution.
\newblock In {\em CVPR}, pages 10581--10590, 2021.

\bibitem{wang2004image}
Zhou Wang, Alan~C. Bovik, Hamid~R. Sheikh, and Eero~P. Simoncelli.
\newblock Image quality assessment: from error visibility to structural similarity.
\newblock {\em IEEE TIP}, 13(4):600--612, 2004.

\bibitem{wang2020joint}
Zihao~W. Wang, Peiqi Duan, Oliver Cossairt, Aggelos Katsaggelos, Tiejun Huang, and Boxin Shi.
\newblock Joint filtering of intensity images and neuromorphic events for high-resolution noise-robust imaging.
\newblock In {\em CVPR}, pages 1609--1619, 2020.

\bibitem{weng2022boosting}
Wenming Weng, Yueyi Zhang, and Zhiwei Xiong.
\newblock Boosting event stream super-resolution with a recurrent neural network.
\newblock In {\em ECCV}, pages 470--488, 2022.

\bibitem{xu2021motion}
Fang Xu, Lei Yu, Bishan Wang, Wen Yang, Gui-Song Xia, Xu Jia, Zhendong Qiao, and Jianzhuang Liu.
\newblock Motion deblurring with real events.
\newblock In {\em ICCV}, pages 2583--2592, 2021.

\bibitem{xu2014inverse}
Li Xu, Xin Tao, and Jiaya Jia.
\newblock Inverse kernels for fast spatial deconvolution.
\newblock In {\em ECCV}, pages 33--48, 2014.

\bibitem{zheng2023survey}
Zheng Xu, Liu Yexin, Lu Yunfan, Hua Tongyan, Pan Tianbo, Zhang Weiming, Tao Dacheng, and Wang Lin.
\newblock Deep learning for event-based vision: A comprehensive survey and benchmarks.
\newblock {\em arXiv preprint arXiv:2302.08890}, 2023.

\bibitem{Zamir2021MPRNet}
Syed~Waqas Zamir, Aditya Arora, Salman Khan, Munawar Hayat, Fahad~Shahbaz Khan, Ming-Hsuan Yang, and Ling Shao.
\newblock Multi-stage progressive image restoration.
\newblock In {\em CVPR}, 2021.

\bibitem{zhang2022unifying}
Xiang Zhang and Lei Yu.
\newblock Unifying motion deblurring and frame interpolation with events.
\newblock In {\em CVPR}, pages 17765--17774, 2022.

\bibitem{zhang2021exposure}
Youjian Zhang, Chaoyue Wang, Stephen~J. Maybank, and Dacheng Tao.
\newblock Exposure trajectory recovery from motion blur.
\newblock {\em IEEE TPAMI}, 44(11):7490--7504, 2021.

\bibitem{zhao2022transformer}
Junwei Zhao, Shiliang Zhang, and Tiejun Huang.
\newblock Transformer-based domain adaptation for event data classification.
\newblock In {\em ICASSP}, pages 4673--4677, 2022.

\bibitem{zihao2018unsupervised}
Alex Zihao~Zhu, Liangzhe Yuan, Kenneth Chaney, and Kostas Daniilidis.
\newblock Unsupervised event-based optical flow using motion compensation.
\newblock In {\em ECCV}, 2018.

\end{thebibliography}

\end{document}